\documentclass{article} 
\usepackage{iclr2019_conference,times}

\usepackage{thmtools}
\usepackage{thm-restate}
\usepackage{hyperref}
\usepackage{wrapfig}

\usepackage[normalem]{ulem}
\usepackage{url}
\usepackage{multirow}
\usepackage{xspace}
\usepackage{amsmath}
\usepackage{cleveref}
\usepackage{amsthm}
\usepackage{textcomp}
\usepackage{amssymb}
\usepackage{amsfonts}
\usepackage[final]{graphicx}
\usepackage{threeparttable}
\usepackage{enumitem}
\usepackage{parskip}
\usepackage{mathrsfs}
\usepackage{booktabs}
\usepackage{caption}
\usepackage{tcolorbox}
\usepackage{times}
\usepackage{graphicx}
\usepackage{bbm}
\usepackage{color}
\usepackage{listings}

\usepackage{tikz}
\usepackage{tkz-graph}
\usetikzlibrary{positioning}
\usepackage{subfig}
\usepackage{booktabs}
\usepackage{makecell}
\usepackage{rotating}
\usepackage{pdflscape}
\usepackage{algorithm}
\usepackage[noend]{algpseudocode}
\usepackage{csquotes}
\usepackage{ulem}

\title{Janossy Pooling: Learning Deep Permutation-invariant Functions for Variable-size Inputs}
\author{Ryan L. Murphy \\
Department of Statistics\\
Purdue University\\
\texttt{murph213@purdue.edu} \\
\And 
Balasubramaniam Srinivasan \\
Department of Computer Science \\
Purdue University \\
\texttt{bsriniv@purdue.edu}
\And
Vinayak Rao \\
Department of Statistics \\
Purdue University \\
\texttt{varao@purdue.edu} \\
\And
Bruno Ribeiro \\
Department of Computer Science \\
Purdue University \\
\texttt{ribeiro@cs.purdue.edu}
}

\newcommand{\Janossy}{Janossy\xspace}

\newcommand{\graphsage}{GraphSAGE\xspace}

\newcommand{\NA}{--}

\newcommand{\tradn}[1]{p_{\mathrm{trad}, n}}

\newcommand{\ftheta}{\boldsymbol{\theta}^{(f)}}
\newcommand{\rhoTheta}{\boldsymbol{\theta}^{(\rho)}}
\newcommand{\htheta}{\boldsymbol{\theta}^{(h)}}
\newcommand{\Jloss}{J}
\newcommand{\vx}{{\bf X}}

\newcommand{\Appendix}{Supplementary Material\xspace}
\newcommand{\pj}[1]{\downarrow_{#1}\!\!}
\newcommand{\deepsets}{DeepSets}
\newcommand*\dbar[1]{\overline{\overline{\lower0.2ex\hbox{$#1$}}}}

\newcommand{\harrow}[1]{\mathstrut\mkern2.5mu#1\mkern-11mu\raise1.6ex%
  \hbox{$\scriptscriptstyle\rightharpoonup$}}

\newcommand{\harrowStable}[1]{\overset{\rightharpoonup}{#1}}

\newcommand{\kcard}{|\vhid|}

\newcommand{\sData}{\sX}
\newcommand{\sHid}{\sH}

\newcommand{\evdata}{\evx}
\newcommand{\vdata}{\vx}
\newcommand{\evhid}{\evh}
\newcommand{\vhid}{\vh}
\newcommand{\thetaToF}{\vtheta^{(f)}}

\colorlet{darkgreen}{green!50!black}

\algnewcommand\algorithmicforeach{\textbf{for each}}
\algdef{S}[FOR]{ForEach}[1]{\algorithmicforeach\ #1\ \algorithmicdo}

\newcommand{\btheta}{\boldsymbol{\theta}}

\newtheorem{proposition}{Proposition}[section]

\newenvironment{restat}
{\restatable{theorem}{staircase}}
{\endrestatable}

\newenvironment{restatProp}
{\restatable{proposition}{pisgd}}
{\endrestatable}

\newenvironment{restatKaryProp}
{\restatable{proposition}{karytract}}
{\endrestatable}

\newtheorem{corollary}{Corollary}[section]
\newtheorem{remark}{Remark}[section]

\newtheorem{definition}{Definition}[section]
\renewenvironment{definition}{\refstepcounter{definition}\par\noindent\textit{Definition~\thedefinition:}\xspace}{\nobreak\hfill$\Diamond$\par}

\usepackage{amsmath,amsfonts,bm}

\def\secref#1{section~\ref{#1}}

\def\eqref#1{equation~\ref{#1}}
\def\eqrefs#1#2{equations~\ref{#1} and~\ref{#2}}
\def\Eqref#1{Equation~\ref{#1}}

\def\1{\bm{1}}
\newcommand{\train}{\mathcal{D}}

\def\rvs{{\mathbf{s}}}

\def\rmZ{{\mathbf{Z}}}

\def\vtheta{{\bm{\theta}}}

\def\vh{{\bm{h}}}

\def\vx{{\bm{x}}}
\def\vy{{\bm{y}}}

\def\evh{{h}}

\def\evx{{x}}

\def\mG{{\bm{G}}}

\DeclareMathAlphabet{\mathsfit}{\encodingdefault}{\sfdefault}{m}{sl}
\SetMathAlphabet{\mathsfit}{bold}{\encodingdefault}{\sfdefault}{bx}{n}

\def\gB{{\mathcal{B}}}

\def\sF{{\mathbb{F}}}

\def\sH{{\mathbb{H}}}
\def\sI{{\mathbb{I}}}

\def\sN{{\mathbb{N}}}

\def\sR{{\mathbb{R}}}

\def\sX{{\mathbb{X}}}
\def\sY{{\mathbb{Y}}}

\newcommand{\expected}{E}

\newcommand{\normltwo}{L^2}

\def\COMPLETE{}
\iclrfinalcopy 
\begin{document}
\maketitle
\begin{abstract}

We consider a simple and overarching representation for permutation-invariant functions of sequences (or {multi}set functions). 
Our approach, which we call \Janossy pooling, expresses a permutation-invariant function as the average of a permutation-sensitive function applied to all reorderings of the input sequence. 
This allows us to leverage the rich and mature literature on permutation-sensitive functions to construct novel and flexible permutation-invariant functions. 
If carried out naively, \Janossy pooling can be computationally prohibitive. 
To allow computational tractability, we consider three kinds of approximations: canonical orderings of sequences, functions with $k$-order interactions, and stochastic optimization algorithms with random permutations. 
Our framework 
unifies a variety of existing work in the literature, and suggests possible modeling and algorithmic extensions. 
We explore a few in our experiments, which demonstrate improved performance over current state-of-the-art methods. 

\end{abstract}


\newcommand{\rhotheta}{\vtheta^{(\rho)}}

\vspace{-.12in}
\section{Introduction}\label{sec:intro}
\vspace{-.1in}
Pooling is a fundamental operation in deep learning architectures~\citep{LeCun2015}. 
The role of pooling is to merge a collection of related features into a single, possibly vector-valued, summary feature. 
A prototypical example is in convolutional neural networks (CNNs)~\citep{lecun1995convolutional}, where linear activations of features in neighborhoods of image locations are pooled together to construct more abstract features.
A more modern example is in neural networks for graphs, where each layer pools together embeddings of neighbors of a vertex to form a new 
embedding for that vertex, see for instance, \citep{Kipf2016, Atwood2016, Hamilton2017, velickovic2017graph, monti2017geometric, Xu2018, Liu2018, liben2007link, van2017graph, Duvenaud2015, Gilmer2017NeuralMP, ying2018hierarchical, xu2018how}.

A common requirement of a pooling operator is invariance to the ordering of the input features. 
In CNNs for images, pooling allows invariance to translations and rotations, while for graphs, it allows invariance to 
graph isomorphisms. 
Existing pooling operators are mostly limited to pre-defined heuristics such as max-pool, min-pool, sum, or average. 
Another desirable characteristic of pooling layers is the ability to take variable-size inputs. This is less important in images, where neighborhoods 
are usually fixed {\em a priori}. However in applications involving graphs, the number of neighbors of different vertices can vary widely. 
Our goal is to design flexible and learnable pooling operators satisfying these two desiderata.

Abstractly, we will view pooling as a permutation-invariant (or symmetric) function acting on finite but arbitrary length sequences $\vhid$. All elements 
$\evhid_i$ of the sequences are features lying in some space $\sHid$ (which itself could be a high-dimensional Euclidean space $\sR^d$ or some subset 
thereof). The sequences $\vhid$ are themselves elements of the union of products of the $\sHid$-space: 
$\vhid \in \bigcup_{j=0}^\infty \sHid^j \equiv \sH^\cup$.
Throughout the paper, we will use $\Pi_n$ to represent the set of all permutations of the integers $1$ to $n$, where $n$ will often be clear from the context. 
In addition, $\vhid_\pi$, $\pi \in \Pi_{|\vhid|}$, will represent a reordering of the elements of a sequence $\vhid$ according to $\pi$, where $|\vhid|$ is the length of the sequence $\vhid$.
We will use the double bar superscript 
$\dbar{f}$ to indicate that a function is permutation-invariant, returning the same value no matter the order of its arguments: $\dbar{f}(\vhid) = \dbar{f}(\vhid_\pi)$, $\forall \pi \in \Pi_{|\vhid|}$. 
We will use the arrow superscript $\harrow{f}$ to indicate general functions on sequences $\vhid$ which may or may not be 
permutation-invariant%
\footnote{\LaTeX  code for these markers is provided\\ in the~\Appendix.}. %
Functions $f$ without any markers are `simple' functions, acting on elements in $\sHid$, scalars or any other argument 
that is not a sequence of elements in $\sHid$. 

Our goal in this paper is to model and learn permutation-sensitive functions $\harrow{f}$ that can be used to construct flexible and 
learnable permutation-invariant neural networks. A recent step in this direction is work on {\em \deepsets} by~\cite{Zaheer2017}, who argued for learning permutation-invariant functions
through the following composition:
\vspace{-.05in}
\begin{align} 
  \dbar{y}(\vdata ; \vtheta^{(\rho)}, \vtheta^{(f)}, \vtheta^{(h)}) &= \rho\left(\dbar{f}(|\vhid|, \vhid; \vtheta^{(f)}); \vtheta^{(\rho)} \right), \text{where} 
\label{eq:deepsets} \\
\dbar{f}(|\vhid|, \vhid; \vtheta^{(f)}) &= 
\sum_{j=1}^{|\vhid|} f(\evhid_j ; \vtheta^{(f)}) \quad \text{ and } \quad \vhid \equiv h(\vdata ; \vtheta^{(h)}).\label{eq:deepsets2}
\vspace{-.1in}
\end{align}
Here, (a) $\vx \in \sData$ is one observation in the training data {($\sData$ itself may contain variable-length sequences)}, $\vh \in \sH$ is the embedding (output) of the data given by the lower layers $h: \sData \times \sR^{a} \to \sH^\cup$, $a > 0$ with parameters $\htheta \in \sR^{a}$;
(b) $f:\sH \times \sR^{b} \to \sF$ is a middle-layer embedding function with parameters $\ftheta \in \sR^b$, $b > 0$, and $\sF$ is the embedding space of $f$;
and 
(c) $\rho: \sF \times \sR^{c} \to \sY$ is a neural network with parameters $\rhotheta \in \sR^{c}$, $c > 0$, that maps to the final output space $\sY$.
Typically $\sH$ and $\sF$ are high-dimensional real-valued spaces; $\sY$ is often $\sR^d$ in $d$-dimensional regression problems or the simplex in classification problems.
Effectively, the neural network $f$ learns an embedding for each element in $\sHid$,
and given a sequence $\vhid$, its component embeddings are added together before a second neural network transformation $\rho$ is applied. Note that the function $h$ may be the identity mapping $h(\vdata; \cdot) = \vdata$ that makes $\dbar{f}$ act directly on the input data.
\cite{Zaheer2017} argue that if $\rho$ is a universal function approximator, the above architecture is capable of approximating any symmetric function on $\vhid$-sequences, which justifies the widespread use of average (sum) pooling to make neural networks permutation-invariant in \citet{Duvenaud2015}, \citet{Hamilton2017}, \citet{Kipf2016}, \citet{Atwood2016}, among other works. 
{We note that~\citet{Zaheer2017} focus on functions of sets 
  but the work was extended to functions of multisets by~\citet{xu2018how} and that Janossy pooling can be used to represent multiset functions.   
}

\begin{wrapfigure}[22]{r}{0.58\textwidth}
	\vspace{-13pt}
	\begin{minipage}[h]{0.54\textwidth}
		\begin{center}
			\includegraphics[scale=0.4]{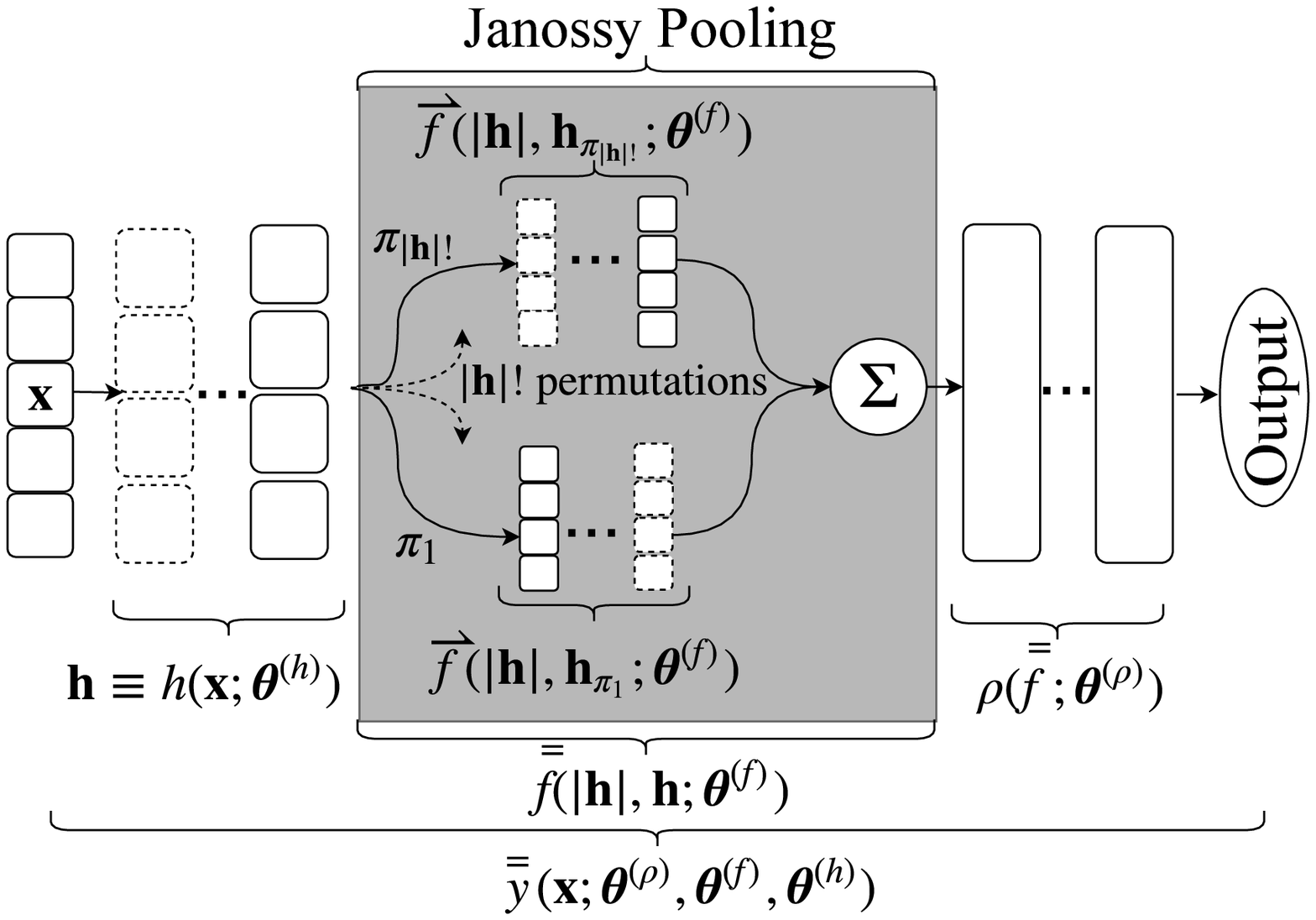}
		\end{center}
	\end{minipage}
	\begin{minipage}[h]{0.54\textwidth}
		\caption{
			{A neural network with a single \Janossy pooling layer. The embedding $\vhid$ is permuted in all $|\vhid|!$ possible ways, and for each permutation  $\vhid_\pi$, $\overset{\rightharpoonup}{f}(|\vhid|, \vhid_\pi ; \ftheta)$ is 
			computed.
			These are 
			summed and 
			passed to a second function 
			$\rho(\cdot; \rhoTheta)$ which gives the final permutation-invariant output $\dbar{y}(\vdata ;  \vtheta^{(\rho)}, \vtheta^{(f)}, \vtheta^{(h)})$; the gray rectangle represents \Janossy pooling.
			We discuss how this can be made computationally tractable.} 
		}
		\label{fig:architecture}
	\end{minipage}

\end{wrapfigure}

In practice, there is a gap between flexibility and learnability. 
While the architecture of~\eqrefs{eq:deepsets}{eq:deepsets2} is a universal approximator to permutation-invariant functions, it does not easily encode structural knowledge about $\dbar{y}$. 
Consider trying to learn the permutation-invariant function $\dbar{y}(\vdata) = \max_{i,j \le |\vdata|} | \evdata_i - \evdata_j|$.
With higher-order interactions between the elements of $\vhid$, the functions $f$ of \eqref{eq:deepsets2} cannot capture any useful intermediate representations towards the final output, 
with the burden shifted entirely to the function $\rho$. 
Learning $\rho$ means learning to undo mixing 
performed by the summation layer $\dbar{f}(|\vhid|, \vhid; \vtheta^{(f)}) = \sum_{j=1}^{|\vhid|} f(\evhid_j ; \vtheta^{(f)})$.
As we show in our experiments, in many applications this is too much to ask of $\rho$.
\clearpage
\paragraph{ \textbf{Contributions.}}
We investigate a learnable permutation-invariant pooling layer for variable-size inputs inspired by the Janossy density 
framework, widely used in the theory of point processes~\citep[Chapter 7]{daley2007introduction}.
This approach, which we call \Janossy pooling, directly allows the user to model what higher-order dependencies in $\vhid$ are relevant in the pooling. 

Figure~\ref{fig:architecture} summarizes a neural network with a  single \Janossy pooling layer $\dbar{f}$ (detailed in Definition~\ref{def:AnySizeDf} below): given an input embedding $\vhid$, we apply a learnable (permutation-sensitive) function $\harrow{f}$ to {every} permutation $\vhid_\pi$ of the input sequence $\vhid$.
These outputs are added together, and fed to the second function $\rho$.
Examples of function $\harrow{f}$ include feedforward and recurrent neural networks (RNNs). 
We call the operation used to construct $\dbar{f}$ from $\harrow{f}$ the {\em \Janossy pooling}.
Definition~\ref{def:AnySizeDf}  gives a more detailed description. {We will detail three broad strategies for making this computation tractable and discuss how existing methods can be seen as tractability strategies under the Janossy pooling framework}.  

Thus, we propose a framework and tractability strategies that unify and extend existing methods in the literature.  We contribute the following analysis:  %
(a) We show {\em \deepsets}~\citep{Zaheer2017} is a special case of \Janossy pooling where the function $\harrow{f}$ depends only on the first element of the {sequence 
} $\vhid_\pi$. In the most general form of \Janossy pooling (as described above), $\harrow{f}$ depends on its entire input sequence $\vhid_\pi$. %
This naturally raises the possibility of intermediate choices of $\harrow{f}$ that allow practitioners to trade between flexibility and tractability.
We will show that functions $\harrow{f}$ that depend on their first $k$ arguments of $\vhid_\pi$ allow the \Janossy pooling layer to capture up to $k$-ary dependencies in $\vhid$. 
(b) We show \Janossy pooling can be used to learn permutation-invariant neural networks $\dbar{y}(\vdata)$ by sampling a random permutation {of $\vh$} during training, and then modeling this permuted sequence using a sequence model such as a recurrent neural network (LSTMs~\citep{hochreiter1997long}, GRUs~\citep{cho2014learning}) or a vector model such as a feedforward network.
We call this permutation-sampling learning algorithm $\pi$-SGD ($\pi$-Stochastic Gradient Descent).
Our analysis explains why this seemingly unsound procedure is theoretically justified, which sheds light on the recent puzzling success of permutation sampling and LSTMs in relational models~\citep{moore2017deep, Hamilton2017}.
We show that this property relates to randomized model ensemble techniques. %
(c) In~\citet{Zaheer2017}, the authors describe a connection between {\em \deepsets} and infinite de Finetti exchangeabilty. We provide a probabilistic connection between \Janossy pooling and {\em finite} de Finetti exchangeabilty~\citep{diaconis1977finite}.
%

\vspace{-.1in}
\section{\Janossy Pooling}
\vspace{-.1in}
\label{sec:JanossyPooling}
We first formalize the \Janossy pooling function $\dbar{f}$.
Start with a function $\harrow{f}$, parameterized by $\ftheta$, which can take any variable-size sequence as input: a sequence of matrices (such as 
images), a sequence of vectors (such as a sequence of vector embeddings), or a variable-size sequence of features or embeddings representing the 
neighbors of a node in an attributed graph. In practice, we implement $\harrow{f}$ with a neural network.
Formalizing Figure~\ref{fig:architecture} from Section~\ref{sec:intro},
we use $\harrow{f}$ to define $\dbar{f}$:

\begin{definition}[\Janossy pooling]
\label{def:AnySizeDf}
Consider a function $\harrow{f}: \sN \times \sH^\cup \times \sR^b \to \sF$ on variable-length but finite sequences $\vh$, parameterized by $\ftheta \in \sR^b$, $b > 0$.
A permutation-invariant function $\dbar{f}: \mathbb{N} \times \sH^\cup \times \sR^b \to \sF$ is the \Janossy function associated with $\harrow{f}$ if
\begin{equation}
\label{eq:JanossyPoolingN}
\dbar{f}(|\vhid|, \vhid ; \ftheta ) = \frac{1}{|\vhid|!} \sum_{\pi \in \Pi_{|\vhid|}} \harrow{f}(|\vhid|, \vhid_{\pi}; \ftheta),
\end{equation}
where $\Pi_{|\vhid|}$ is the set of all permutations of the integers $1$ to $|\vhid|$, and $\vhid_\pi$ represents a particular reordering of the elements of sequence $\vhid$ according to $\pi \in \Pi_{|\vhid|}$.
We refer the operation used to construct $\dbar{f}$ from $\harrow{f}$ as \Janossy pooling.
\end{definition}

Definition~\ref{def:AnySizeDf} provides a conceptually simple approach 
for constructing permutation-invariant functions from arbitrary and powerful
permutation-sensitive functions such as feedforward networks, recurrent neural networks, or convolutional neural networks.  If $\harrow{f}$ is a vector-valued function, then so is $\dbar{f}$, and in practice, one might pass this 
vector output of $\dbar{f}$ through a second function $\rho$ (e.g.\ a neural network parameterized by $\theta^{(\rho)}$): 
\begin{equation}
\label{eq:JanossyPoolingRho}
\dbar{y}(\vdata ; \vtheta^{(\rho)}, \vtheta^{(f)}, \vtheta^{(h)}) = \rho\left(\frac{1}{|\vhid|!} \sum_{\pi \in \Pi_{|\vhid|}} \harrow{f}(|\vhid|, \vhid_{\pi}; \ftheta); \vtheta^{(\rho)}\right), \text{where}  \quad \vhid \equiv h(\vdata ; \vtheta^{(h)}).
\end{equation}
\Eqref{eq:JanossyPoolingN} can capture any permutation-invariant function $\dbar{g}$ for a flexible enough family of permutation-sensitive functions $\harrow{f}$ (for instance, one could always set $\harrow{f} = \dbar{g}$). Thus, at least theoretically, $\rho$ in \eqref{eq:JanossyPoolingRho} provides no additional representational power. In practice, however, $\rho$ can improve learnability by capturing common aspects across all terms in the summation. 
Furthermore, when we look at approximations to~\eqref{eq:JanossyPoolingN} or 
restrictions of $\harrow{f}$ to more tractable families, adding $\rho$ can help recover some of the lost 
model capacity. Overall then,~\eqref{eq:JanossyPoolingRho} represents one layer of \Janossy pooling, forming a constituent part of a bigger neural 
network. Figure~\ref{fig:architecture} summarizes this.

\Janossy pooling, as defined in~\eqref{eq:JanossyPoolingN} and \ref{eq:JanossyPoolingRho} is  intractable; the computational cost of summing over all 
permutations (for prediction), and backpropagating gradients (for learning) is likely prohibitive for most problems of interest. 
Nevertheless, it provides an overarching framework to unify existing methods, 
and to extend them. In what follows we present strategies for mitigating this, 
allowing novel and effective trade-offs between 
learnability and computational cost.

\subsection{Tractability through Canonical Input Orderings} \label{sec:tract_canon}
A simple way to achieve permutation-invariance without the summation in \eqref{eq:JanossyPoolingN} is to order the elements of $\vhid$ according to some canonical ordering based on its values, and then feed the reordered sequence to $\harrow{f}$. More precisely, one defines a function $\mathrm{CANONICAL} : \sH^\cup \to \sH^\cup $ such that $\mathrm{CANONICAL}(\vhid) =\mathrm{CANONICAL}(\vhid_{\pi}) \forall \pi \in \Pi_{|\vhid|}$ and only considers functions $\harrow{f}$ based on the composition $\harrow{f} = \mathrm{CANONICAL} \circ \harrow{f}^{\prime}$.  Note that specifying a permutation-invariant $\mathrm{CANONICAL}$ is not equivalent to the original problem since one may define a function of only the data and not of learnable parameters (e.g. sort). 
This input constraint then allows the use of complex $\harrow{f}$ models, such as RNNs, that can capture arbitrary relationships in the canonical ordering of $\vhid$ without the need to sum over all permutations of the input.

Examples of the canonical ordering approach already exist in the literature, 
for example,~\citet{niepert2016learning} order nodes in a graph according to 
a user-specified ranking such as betweenness centrality (say from high to low). 
This approach is useful only if the canonical ordering is relevant to the task at hand. \citet{niepert2016learning}\! acknowledges this 
shortcoming and \citet{moore2017deep} demonstrates that an ordering by Personalized PageRank  \citep{Page_Brin_Motwani_Winograd_1999, Jeh_Widom_2003} achieves a lower classification accuracy than a random ordering.
As an idealized example, consider input sequences $\vhid = \big( (\evhid_{i,1}, \evhid_{i,2}) \big)_{i=1}^{n}$, with $(\evhid_{i,1}, 
\evhid_{i,2}) \in \sHid = \sR^2$, and components $\evhid_{i,1}$ and $\evhid_{i,2}$ sampled independently of each other.
Choosing to sort $\vhid$ according to $\evhid_{\cdot,1}$ when the task at hand depends on sorting according to $\evhid_{\cdot,2}$ can lead to poor prediction accuracy.

Rather than pre-defining a good canonical order, one can try to learn it from the data. This requires searching over the discrete space of all $|\vh|!$ permutations of the input vector $\vh$. In practice, this discrete optimization relies on heuristics~\citep{Vinyals2016,rezatofighi2018deep}. 
Alternatively, instead of choosing a single canonical ordering, one can choose multiple orderings, resulting in ensemble methods that average across multiple 
permutations. These can be viewed as more refined (possibly data-driven) approximations to~\eqref{eq:JanossyPoolingN}.

\subsection{Tractability through $k$-ary dependencies}
\label{sec:kAry}

Here, we provide a different spectrum of options to trade-off flexibility, complexity, and generalizability in \Janossy pooling.
Now, to simplify the sum over permutations in \eqref{eq:JanossyPoolingN}, we impose structural constraints where $\harrow{f}(\vhid)$ depends only on 
the first $k$ elements of its input sequence. This amounts to the assumption that only $k$-ary dependencies in $\vhid$ are relevant to the task at hand. 
\begin{definition}[$k$-ary \Janossy pooling] \label{def:GeneralPool}%
Fix $k \in \mathbb{N}$.  
For any sequence $\vhid$, define $\pj{k}(\vhid)$ as its projection to a length 
$k$ sequence; in particular, if $|\vhid| \ge k$, we keep the first $k$ elements. 
Then, a $k$-ary permutation-invariant Janossy function $\dbar{f}$ is given by 
\begin{equation} \label{eq:fKary}
   \dbar{f}(|\vhid|, \vhid ; \ftheta ) = \frac{1}{|\vhid|!} \sum_{\pi \in \Pi_{|\vhid|}} \harrow{f}(\kcard, \pj{k}(\vhid_{\pi}); \ftheta).
\vspace{-.2in}
\end{equation}
\end{definition}
%
%
{Note that if some of the embeddings have length $|\vhid| < k$, then we 
can zero pad to form the length-$k$ sequence $(\pj{k}(\vhid_{\pi}), 0, \ldots, 0)$}. 
Proposition~\ref{prop:kPerms} shows that if $|\vhid| > k$, \eqref{eq:fKary} only needs to sum over $|\vhid|!/(|\vhid|-k)!$ terms,  
which can be tractable for small $k$.
\begin{restatKaryProp}
\label{prop:kPerms}
The \Janossy pooling in \eqref{eq:fKary} requires summing over only 
$\frac{|\vhid|!}{(|\vhid|-k)!}$ terms, thus saving computation when $k < |\vhid|$. In particular,~\eqref{eq:fKary} can be written as $\frac{(|\vhid|-k)!}{|\vhid|!}\sum_{(i_1, i_2, \ldots, i_k) \in \sI_{|\vhid|}} \harrow{f}\big(\kcard, (\evhid_{i_1}, \evhid_{i_2}, \ldots, \evhid_{i_k}) ; \ftheta \big),$ where $\sI_{|\vhid|}$ is the set of all permutations of $\{1, 2, \ldots, |\vhid| \}$ taken $k$ at a time, and $\evhid_{j}$ is the $j$-th element of $\vh$.
\end{restatKaryProp}
Note that the value of $k$ balances computational savings and the capacity to model higher-order interactions; it can be selected as a hyperparameter based on a-priori beliefs or through typical hyperparameter tuning strategies.  
\begin{remark}[\deepsets~\citep{Zaheer2017} is a $1$-ary (unary) \Janossy pooling]
\label{rmk:EncapsulateDeepSet} 
\Eqref{eq:fKary} represented with $k=1$ and composing with $\rho$ as in~\eqref{eq:JanossyPoolingRho} yields the model 
$\rho\big(\frac{1}{|\vhid|}\sum_{j=1}^{|\vhid|} \harrow{f}(\kcard, \evhid_j;\ftheta); \rhotheta \big)$  and thus equations~\ref{eq:deepsets} and \ref{eq:deepsets2} for an appropriate choice of $\harrow{f}$.
\end{remark}

%
%
Not surprisingly, the computational savings obtained from $k$-ary 
\Janossy pooling come at the cost of reduced model flexibility. 
The next result formalizes this.

\begin{restat}
\label{thm:k_implies_km}
For any $k\in\mathbb{N}$, define $\mathcal{F}_{k}$ as the set of all 
permutation-invariant functions that can be represented
by \Janossy pooling with $k$-ary dependencies.
Then, $\mathcal{F}_{k-1}$ is a \emph{proper} subset of $\mathcal{F}_k$ if the space $\sHid$ is not trivial (i.e.\ if 
the cardinality of $\sHid$ is greater than $1$).
Thus, \Janossy pooling with $k$-ary
dependencies can express any \Janossy pooling function with $(k-1)$-ary
dependencies, but the converse does not hold.
\vspace{-5pt}
\end{restat}

The proof is given in the~\Appendix. Theorem~\ref{thm:k_implies_km} has the following implication:
%
\begin{corollary}
\label{cor:StaircaseAndDeepSets}
For $k > 1$, the {\em \deepsets} function in \eqref{eq:deepsets}~\citep{Zaheer2017} pushes the modeling of $k$-ary relationships to $\rho$.
\vspace{-9pt}
\end{corollary}
\begin{proof}
{\em \deepsets} functions can 
be expressed via Janossy pooling with $k=1$. 
Thus, by Theorem~\ref{thm:k_implies_km}, $\dbar{f}$ in \eqref{eq:deepsets2} cannot express all functions
that can be expressed by higher-order (i.e. $k>1$) Janossy pooling
operations. Consequently, if the {\em \deepsets} function can express any permutation-invariant function, the expressive power must have been pushed to $\rho$.  
\end{proof}

%
%
\newcommand{\vpi}{\pi}
\newcommand{\evpi}{\pi}
\newcommand{\sPi}{\Pi}

\subsection{Tractability through Permutation Sampling}
\label{sec:PermSampling}
Another approach to tractable \Janossy pooling samples {\em random} permutations of the input $\vh$ during training. 
Like the {canonical ordering} approach of Section~\ref{sec:tract_canon}, this offers significant computational savings,
allowing more complex models for $\harrow{f}$ such as LSTMs and GRUs. However, in contrast with that approach, this is 
considerably more flexible, avoiding the need to learn a canonical ordering or to make assumptions about the dependencies between the elements of $\vhid$ and the objective function.
Rather, it can be viewed as {\em implicitly} assuming simpler structure in these functions.
The approach of sampling random permutations has been previously used in relational learning tasks~\citep{moore2017deep,Hamilton2017} as a heuristic 
with an LSTM as $\harrow{f}$. Both these papers report that permutation sampling outperforms or closely matches other tested neural network models they tried.
Therefore, this section not only proposes a tractable approximation for~\eqref{eq:JanossyPoolingN} but also provides a theoretical framework to understand and extend such approaches. 

For the sake of simplicity, we analyze the optimization with a single sampled permutation.  However, note that increasing the number of sampled permutations in the estimate of $\dbar{f}$ decreases variance, and we recover the exact algorithm when all $|\vh|!$ permutations are sampled.
We assume a supervised learning setting, though our analysis easily extends to unsupervised learning. 
We are given training data $\train \equiv\{(\vdata(1),\vy(1)),\ldots,(\vdata(N),\vy(N))\}$, where $\vy(i) \in \sY$ is the target output and $\vdata(i)$ its 
corresponding input. 
%
Our original goal was to minimize the empirical loss
\begin{equation}\label{eq:Loss}
 \vspace{-.1in}
\dbar{L}(\train;\rhotheta,\ftheta,\htheta)  = \frac{1}{N}\sum_{i=1}^N \! L\left(\vy(i) , 
\rho\Big( \dbar{f}(|\vhid^{(i)}|, \vhid^{(i)};\ftheta) ; \vtheta^{(\rho)} \Big) \right) , \text{where} 
\end{equation}
\begin{equation}\label{eq:janLoss}
\dbar{f}(|\vhid^{(i)}|, \vhid^{(i)};\ftheta) =  \frac{1}{|\vhid^{(i)}|!} \sum_{\pi \in \Pi_{|\vhid^{(i)}|}} \!\!\harrow{f}(|\vhid^{(i)}|, \vhid^{(i)}_\pi; \ftheta)
\end{equation}
and $\vhid^{(i)} = h(\vdata(i); \vtheta^{(h)}) \in  \sH^\cup $ with $\vhid^{(i)}_\pi \equiv  (\vhid^{(i)})_\pi$ for permutation $\pi$.
Computing the gradient of \eqref{eq:Loss} is intractable for large inputs $\vh^{(i)}$, as the backpropagation computation graph branches out for every permutation in the sum. To address this computational challenge, we will turn our attention to stochastic optimization.

{\bf Permutation sampling.} Consider replacing the \Janossy sum in \eqref{eq:janLoss}  with the estimate
\begin{equation}
\hat{\dbar{f}}(|\vhid|, \vhid ; \ftheta ) = \harrow{f}(|\vhid|, \vhid_{\rvs}; \ftheta),
\label{eq:JanStochOpt}
\end{equation}
where $\rvs$ is a random permutation sampled uniformly, $\rvs \sim \text{Unif}(\Pi_{|\vhid|})$.
The estimator in \eqref{eq:JanStochOpt} is unbiased: $E_\rvs[\hat{\dbar{f}}(|\vhid|, \vhid_{\rvs} ; \ftheta )] = \dbar{f}(|\vhid|, \vhid ; \ftheta ).$
Note however that when $\dbar{f}$ is chained with another nonlinear 
function $\rho$ and/or nonlinear loss $L$, the composition is no longer unbiased:
$E_\rvs[L(\vy,\rho(\harrow{f}(|\vh_\rvs|, \vh_\rvs; \vtheta^{(f)});\vtheta^{(\rho)}))] \neq 
L(\vy,\rho(E_\rvs[\harrow{f}(|\vh_\rvs|, \vh_\rvs; \vtheta^{(f)})];\vtheta^{(\rho)})) $. 
Nevertheless, we use this estimate to propose the following stochastic approximation algorithm for gradient descent:
\begin{definition}[$\pi$-SGD]\label{d:piSGD}
Let $\gB =\{(\vdata(1),\vy(1)),\ldots,(\vdata(B),\vy(B))\}$ be a mini-batch i.i.d.\ sampled uniformly from the training data $\train$.
At step $t$, consider the stochastic gradient descent update 
\begin{equation}\label{eq:piSGD}
\vtheta_t = \vtheta_{t-1} - \eta_{t} \rmZ_t,
\end{equation}
where 
$
\rmZ_t = \frac{1}{B} \sum_{i = 1}^{B} \nabla_{\vtheta} L\left(\vy(i), \rho\Big( \harrow{f}(|\vhid^{(i,t-1)}|, \vhid^{(i,t-1)}_{\rvs_i}; \vtheta^{(f)}_{t-1}) ; \vtheta^{(\rho)}_{t-1} \Big) \right)
$ is the random gradient,
where $\vhid^{(i,t-1)}_\pi \equiv (h(\vdata(i); \vtheta_{t-1}^{(h)}))_\pi$ for a permutation $\pi$, $\vtheta \equiv (\vtheta^{(\rho)}, \vtheta^{(f)}, \vtheta^{(h)})$, with the random permutations $\{\rvs_i\}_{i=1}^B$, {sampled independently} $\rvs_i \sim \text{Uniform}(\Pi_{|\vhid^{(i)}|})$; the learning rate is $\eta_t \in (0,1)$ s.t.\ $\lim_{t\to \infty} \eta_t = 0$, $\sum_{t=1}^\infty \eta_t  = \infty$, and $\sum_{t=1}^\infty \eta_t^2  < \infty$.  
\end{definition}
Effectively, this is a Robbins-Monro stochastic approximation algorithm of gradient descent \citep{Robbins1951,bottou2004large} and optimizes the following modified objective: \begin{equation} \label{eq:RLoss}
\begin{split}
\dbar{\Jloss}(\train;\vtheta^{(\rho)}, \vtheta^{(f)}, \vtheta^{(h)})  &= \frac{1}{N}\sum_{i=1}^N E_{\rvs_i}\left[ 
 \! L\Bigg(\vy(i) , 
\rho\Big( \harrow{f}(|\vhid^{(i)}|, \vhid^{(i)}_{\rvs_i}; \vtheta^{(f)}) ; \vtheta^{(\rho)} \Big) \Bigg)
\right]\\
&= \frac{1}{N}\sum_{i=1}^N  \frac{1}{|\vhid^{(i)}|!} \sum_{\pi \in \sPi_{|\vhid^{(i)}|}} 
 L\Bigg(\vy(i) , 
\rho\Big( \harrow{f}(|\vhid^{(i)}|, \vhid^{(i)}_\pi; \vtheta^{(f)}) ; \vtheta^{(\rho)} \Big) \Bigg).
\end{split}
\end{equation}
Observe that the expectation over permutations is now outside the $L$ and $\rho$ functions.
Like \eqref{eq:Loss}, the loss in \eqref{eq:RLoss} is also permutation-invariant,  though we note that $\pi$-SGD, after a finite number of 
  iterations, returns a $\rho(\harrow{f}(\cdots,\vh^{(i)},\cdots))$ sensitive to the random input permutations of $\vh^{(i)}$ presented to the algorithm.
  Further, unless the function $\harrow{f}$ itself is 
  permutation-invariant ($\dbar{f} = \harrow{f}$), the optima of 
  $\dbar{\Jloss}$ are different from those of the original objective function 
  $\dbar{L}$.
Instead, $\dbar{\Jloss}$ is an upper bound to $\dbar{L}$ via Jensen's inequality if $L$ is convex and $\rho$ is the identity function (\eqref{eq:JanossyPoolingN}); minimizing this upper bound forms a tractable surrogate to the original Janossy objective. 
{If the function class used to model $\harrow{f}$ is rich enough to include permutation-invariant 
  functions, then the global minima of $\dbar{J}$ will include those of $\dbar{L}$. In general,
minimizing the upper bound implicitly regularizes $\harrow{f}$ to return 
functions that are insensitive to permutations of the training data.}
While a general $\rho$ no longer upper bounds the original 
objective, the implicit regularization of permutation-sensitive functions 
still applies to the composition $\harrow{f}^{\prime} \equiv \rho \circ \harrow{f}$ and 
we show competitive results. %

{It is important to observe that the function $\rho$  plays a very different role in our $\pi$-SGD formulation compared to $k$-ary \Janossy pooling.  Previously $\rho$ was composed with an average over $\harrow{f}$ to model dependencies not captured in the average-- and was in some sense separate from $\harrow{f}$ -- whereas here it becomes absorbed directly into $\harrow{f}^{\prime} = \rho \circ \harrow{f}$.} 

The next result, which we state and prove more formally in the~\Appendix, provides some insight into the convergence properties of our algorithm. Although the conditions are difficult to check, they are similar to those used to demonstrate the convergence of SGD, which has been empirically demonstrated to yield strong performance in practice.
\begin{restatProp}[$\pi$-SGD Convergence]\label{p:piSGD}
	The optimization of $\pi$-SGD enjoys properties of almost sure convergence to the optimal $\vtheta$ under similar conditions as SGD.
\end{restatProp}

\paragraph{Variance reduction.}%
Variance reduction of the output of a sampled permutation $\harrow{f}(|\vhid|, \vhid_{\rvs};\ftheta)$, $\rvs \sim \text{Unif}(\Pi_{|\vhid|})$, allows 
$E_\rvs[L(\vy,\harrow{f}(|\vh_\rvs|, \vh_\rvs; \ftheta)] \approx
L(\vy,E_\rvs[\harrow{f}(|\vh_\rvs|, \vh_\rvs; \ftheta)]),$
inducing a near-equivalence between optimizing \eqref{eq:Loss} and \eqref{eq:RLoss}.
Possible approaches include {\em importance sampling} (used by~\citet{chen2018fastgcn} for $1$-ary Janossy), {\em control variates} (also used by \citet{chen2018stochastic} also used for $1$-ary Janossy), Rao-Blackwellization~\cite[Section 1.7]{lehmann2006theory}, and an output regularization, which includes a penalty for two distinct sampled permutations $\rvs$ and $\rvs'$, $\Vert \harrow{f}(|\vhid|, \vhid_{\rvs};\ftheta)  - \harrow{f}(|\vhid|, \vhid_{\rvs'};\ftheta) \Vert_2^2$, so as to reduce the variance of the sampled \Janossy pooling output (used before to improve Dropout masks by~\citet{zolna2017fraternal}).
\vspace{-8pt}
%

\paragraph{Inference.} The use of $\pi$-SGD to optimize the \Janossy pooling layer optimizes the objective $\dbar{\Jloss}$, and thus  has the following 
implication on how outputs should be calculated at inference time:
\begin{remark}[Inference]\label{r:test}
Assume $L(\vy, \hat{\vy})$ is convex as a function of $\hat{\vy}$ (e.g., $L$ is the $\normltwo$ norm, cross-entropy, or negative log-likelihood losses).  
  At test time we estimate the output $\vy(i)$ of input $\vx(i)$ by computing (or estimating)
\begin{equation}\label{eq:yHat}
\hat{\vy}(\vx(i)) = E_{\rvs_i}\left[
\harrow{f'}(|\vhid^{(i,\star)}|, \vhid^{(i,\star)}_{\rvs_i}; \vtheta^{(f')\star})\right] = %
\frac{1}{|\vhid^{(i,\star)}|}\sum_{\pi \in \Pi_{|\vhid^{(i,\star)}|}}\harrow{f'}(|\vhid^{(i,\star)}|, \vhid^{(i,\star)}_{\pi}; \vtheta^{(f')\star}),
\end{equation}
where $\harrow{f'} \equiv \rho \circ \harrow{f}$, $\vtheta^{(f')\star} \equiv (\vtheta^{(f)\star},\vtheta^{(\rho)\star})$,  $\vhid^{(i,\star)}_{\rvs_i} \equiv (h(\vdata(i); \vtheta^{(h)\star}))_{\rvs_i}$ and $\vtheta^{(\rho)\star}, \vtheta^{(f)\star}, \vtheta^{(h)\star}$ are fixed points of the $\pi$-SGD optimization.
\Eqref{eq:yHat} is a permutation-invariant function.
\end{remark}

\paragraph{Combining $\pi$-SGD and \Janossy with $k$-ary Dependencies.}%
In some cases one may consider $k$-ary \Janossy pooling with a moderately large value of $k$ in which case even the summation over $\frac{|\vhid|!}{(|\vhid|-k)!}$ terms (see proposition~\ref{prop:kPerms}) becomes expensive.  In these cases, one may sample $\rvs \sim \text{Unif}\big(\Pi_{|\vhid|} \big)$ and compute $\hat{\dbar{f}}_{k} = \harrow{f}(|\vhid|, \pj{k}(\vhid_{\rvs}); \ftheta)$ in lieu of the sum in~\eqref{eq:fKary}.  Note that \eqref{eq:fKary} defining $k$-ary \Janossy pooling constitutes exact inference of a simplified model whereas $\pi$-SGD with $k$-ary dependencies constitutes approximate inference.  We will return to this idea in our results section where we note that the \graphsage model of \citet{Hamilton2017} can be cast as a $\pi$-SGD approximation of $k$-ary \Janossy pooling.

\newcommand{\balasrini}{balasrini33}
\vspace{-5pt}
\section{Experiments}
\label{sec:results}
\vspace{-5pt}
In what follows we empirically evaluate two tractable \Janossy pooling approaches, $k$-ary dependencies (\secref{sec:kAry}) and sampling permutations for 
stochastic optimization (\secref{sec:PermSampling}), to learn permutation-invariant functions for tasks of different complexities. 
One baseline we compare against is \deepsets~\citep{Zaheer2017}; recall that this corresponds to unary ($k=1$) \Janossy pooling 
(Remark~\ref{rmk:EncapsulateDeepSet}). Corollary~\ref{cor:StaircaseAndDeepSets} shows that explicitly modeling higher-order dependencies during pooling simplifies the 
task of the upper layers ($\rho$) of the neural network, and we evaluate this experimentally by letting $k = 1,2,3,|\vhid|$ over different arithmetic tasks.  
We also evaluate \Janossy pooling in graph tasks, where it can be used as a permutation-invariant function to aggregate the features and embeddings of the 
neighbors of a vertex in the graph. Note that in graph tasks, permutation-invariance is required to ensure that the neural network is invariant to 
permutations in the adjacency matrix (graph isomorphism).  The code used to generate the results in this section are available on GitHub\footnote{\small \url{https://github.com/PurdueMINDS/JanossyPooling}}.

\subsection{Arithmetic Tasks on Sequences of Integers}
\vspace{-5pt}

We first consider the task of predicting the {\em sum} of a sequence of integers~\citep{Zaheer2017} and extend it to predicting other permutation-invariant functions: 
{\em range}, {\em unique sum}, {\em unique count}, and {\em variance}.
In the {\em sum} task we predict the sum of a sequence of 5 integers drawn uniformly at random with replacement from $\{0, 1, \ldots, 99\}$; %
the {\em range} task also receives a sequence 5 integers 
distributed the same way and tries to predict the range (the difference between the maximum and minimum values); %
the  {\em unique sum} task receives a sequence of 10 integers, sampled uniformly with replacement from $\{0, 1, \ldots, 9\}$, and predicts the sum of all unique elements; %
the {\em unique count} task also receives a sequence of repeating elements from $\{0, 1, \ldots, 9\}$, distributed in the same was as with the {\em unique sum} task, and predicts the number of unique elements; %
the {\em variance} task receives a sequence of 10 integers drawn uniformly with replacement from $\{0, 1, \ldots, 99\}$ and tries to predict the variance $ \frac{1}{|\vx|}\sum_{i} (\evx_i - \bar{\vx})^{2} = \frac{1}{2|\vx|^2} \sum_{i,j} (\evx_i - \evx_j)^{2}$, where $\bar{\vx}$ denotes the mean of $\vx$.
Unlike \citet{Zaheer2017}, we choose to work with the digits themselves, to allow a more direct assessment of the 
different \Janossy pooling approximations.
Note that the summation task of \citet{Zaheer2017} is naturally a unary task that lends itself to the approach of embedding individual digits then adding them together while
the other tasks require exploiting high-order relationships within the sequence.  Following~\citet{Zaheer2017}, we report accuracy (0-1 loss) for all tasks with an integer target; we report root mean squared error (RMSE) for the variance task.

Here we explore two \Janossy pooling tractable approximations: \\
(a) ($k$-ary dependencies) {\em \Janossy ($k=1$) (\deepsets)}, and {\em \Janossy $k=2,3$} where $\harrow{f}$ is a feedforward network with a single hidden layer comprised of 30 neurons.  As detailed in the \Appendix, the models are constructed to have the same number of parameters regardless of $k$ by modifying the embedding (output) dimension of $h$. In the \Appendix, we also show results for experiments that relax this constraint.  \\
(b) ($\pi$-SGD) Full $k=|\vh|$ \Janossy pooling where $\harrow{f}$ is an LSTM or a GRU that returns the short-term hidden state of the last temporal unit (the $\vh_{t}$ of \citeauthor{cho2014learning}\! with $t=|\vh|$). The LSTM has 50  hidden units and the GRU 80, trained with the $\pi$-SGD stochastic optimization.  
The number of hidden units was chosen to be consistent with~\citet{Zaheer2017}.
At test time, we experiment with approximating (estimating)~\eqref{eq:yHat} using 1 and 20 sampled permutations.

We also explore two functions for $\rho$ of~\eqref{eq:JanossyPoolingRho} (upper-layer):
(i) [Linear] a single dense layer with identity activation as in the experiments of~\citet{Zaheer2017}, and 
(ii) [MLP (100)] a feedforward network with one hidden layer using tanh activations and 100 units.  Choosing a simple and complex form for $\rho$ allows insight into the extent to which $\rho$ supplements the capacity of the model by capturing relationships not exploited during pooling, and serves as an evaluation of the strategy of optimizing $\dbar{J}$ as a tractable approximation of $\dbar{L}$.

Much of our implementation, architectural, and experimental design are based on the \deepsets   \      code\footnote{\footnotesize \url{https://github.com/manzilzaheer/DeepSets}} of  \citet{Zaheer2017}, see 
the \Appendix for details. We tuned the Adam learning rate for each model and report the results using the rate yielding top performance on the validation set.  Table~\ref{tab:accuracy} shows the accuracy (average 0-1 loss) of all tasks except variance, for which we report RMSE in the last column. Performance was similar between the LSTM and GRU models, with the GRU performing slightly better, thus we moved the LSTM results to Table~\ref{tab:accuracyFull} in the \Appendix for the sake of clarity. We trained each model with 15 random initializations of the weights to quantify variability.  Table~\ref{tab:mae} in the \Appendix  shows the same results measured by mean absolute error. 
The data consists of 100,000 training examples and 10,000 test examples. %

The results in Table~\ref{tab:accuracy} and Table~\ref{tab:accuracyFull} show that: (1) models trained with $\pi$-SGD using LSTMs and GRUs as $\harrow{f}$ typically achieve top performance or are comparable to the top performer (within confidence intervals) on all tasks, for any choice of $\rho$.  We also observe for LSTMs and GRUs that adding complexity to $\rho$ can yield small but meaningful performance gains or maintain similar performance, lending credence to the approach of optimizing $\dbar{J}$ as a tractable approximation to $\dbar{L}$.  (2) Specifically, in the {\em variance} task, GRUs and LSTMs with $\pi$-SGD provide significant accuracy gains over $k \in \{1,2,3\}$, showing that modeling full-dependencies can be advantageous even if model training with $\pi$-SGD is approximate. (3) For a more complex $\rho$ (MLP as opposed to Linear), lower-complexity  \Janossy pooling achieves consistently better results: $k \in \{2,3\}$ gives good results when $\rho$ is linear but poorer results when $\rho$ is an MLP (as these models are more expressive, the only feasible explanation is an optimization issue since we also observed poorer performance on the {\em training data}). We also note that when $\rho$ is an MLP, it takes significantly more epochs for $k \in \{2,3\}$ to find the best model (2000 epochs) while $k=1$ finds good models much quicker (1000 epochs).  	The results we report come from training with 1000 epochs on all models with a linear $\rho$ and 2000 epochs for all models where $\rho$ is an MLP.
(4) We observe that for $k=1$ (\deepsets), a more complex $\rho$ (MLP) is required as the pooling pushes the complexity of modeling high-order interactions over the input  to $\rho$. The converse is also true, if $\rho$ is simple (Linear) then a Janossy pooling that models high-order interactions $k \in \{2,3,|\vh|\}$ gives higher accuracy, as shown in the {\em range}, {\em unique sum}, {\em unique count}, and {\em variance} tasks.

{
\begin{table}[t]
  \small
\vspace{-5pt}
\centering
\caption{Accuracy (and RMSE for the {\em variance} task) of various \Janossy pooling approximations under distinct tasks. The {\em method} column refers to the method used to deal with the sum over all permutations. {\em Infr sample} refers to the number of permutations sampled at test time to estimate \eqref{eq:yHat} for methods learned with $\pi$-SGD. $k=1$ corresponds to \deepsets. $\tanh$ activations are used with the MLP. Standard deviations computed over 15 runs are shown in parentheses.
}
\label{tab:accuracy}
\vspace{-5pt}
\scalebox{0.98}{
\tabcolsep=0.11cm
\begin{tabular}{llrclllllr}
\multicolumn{1}{c}{\em $\harrow{f}$} &\multicolumn{1}{p{1cm}}{\em \centering method} &\multicolumn{1}{p{0.8cm}}{\em \centering infr \\ sample} &\multicolumn{1}{c}{\em k} &\multicolumn{1}{c}{\em $\rho$} &\multicolumn{1}{c}{\em sum} &\multicolumn{1}{c}{\em range} &{\em \centering unique sum} &{\em \centering uniq.\ count}&{\em \centering variance}
\\ \hline
MLP (30) & exact & \NA & \centering{1} & Linear          & 1.00(0.00)     & 0.04(0.00) & 0.07(0.00) & 0.36(0.01) & 119.05(1.29) \\
MLP (30) & exact & \NA & \centering{2} & Linear          & 0.99(0.00) & 0.09(0.00) & 0.17(0.00) & 0.74(0.03)     & 4.37(0.50)     \\
MLP (30) & exact & \NA & \centering{3} & Linear          & 0.99(0.00)  & 0.21(0.00) & 0.44(0.02) & 0.89(0.04)  & 8.99(0.99)    \\
MLP (30) & exact & \NA & \centering{1} & MLP (100) & 1.00(0.00)   & 0.97(0.01) & 1.00(0.00)     & 1.00(0.00)     & 1.95(0.24)      \\
MLP (30) & exact & \NA & \centering{2} & MLP (100) & 1.00(0.00) & 0.97(0.01) & 1.00(0.00)     & 1.00(0.00)     & 3.49(0.48)  \\
MLP (30) & exact & \NA & \centering{3} & MLP (100) & 0.93(0.02) & 0.93(0.02) & 1.00(0.00)     & 1.00(0.00)     & 6.90(0.47) \\
GRU(80) & $\pi$-SGD & 1 & \centering{$|\vh|$} & Linear          & 0.99(0.01) & 0.98(0.00) & 1.00(0.00)     & 1.00(0.00)     & 1.43(0.23)      \\
GRU(80) & $\pi$-SGD & 20 & \centering{$|\vh|$} & Linear          & 0.99(0.00) & 0.99(0.00) & 1.00(0.00)     & 1.00(0.00)     & 1.20(0.23)   \\   
GRU(80) & $\pi$-SGD & 1 & \centering{$|\vh|$} & MLP (100) & 0.99(0.00) & 1.00(0.00) & 1.00(0.00)     & 1.00(0.00)     & 0.42(0.62)    \\
GRU(80) & $\pi$-SGD & 20 & \centering{$|\vh|$} & MLP (100) & 0.99(0.00) & 1.00(0.00) & 1.00(0.00)     & 1.00(0.00)     & 0.40(0.37)     
\end{tabular}
}
\vspace{-10pt}
\end{table}
}

%
%
%
\subsection{\Janossy pooling as an aggregator function for vertex classification}
\label{subsec:graphtasks}

Here we consider Janossy pooling in the context of graph neural networks to learn vertex representations enabling vertex classification.  The \graphsage algorithm \citep{Hamilton2017} consists of sampling vertex attributes from the neighbor multiset of each vertex $v$ before performing an aggregation operation which generates an embedding of $v$; the authors consider permutation-invariant operations such as mean and max as well as the permutation-sensitive operation of feeding a randomly permuted neighborhood sequence to an LSTM.  The sample and aggregate procedure is repeated twice to generate an embedding.  Each step can be considered as Janossy pooling with $\pi$-SGD and $k$-ary subsequences, where $k_l$, $l\in \{1,2\}$ is the number of vertices sampled from each neighborhood and $\harrow{f}$ is for instance a mean, max, or LSTM. However, at test time, \graphsage only samples one permutation $\rvs$ of each neighborhood to estimate~\eqref{eq:yHat}.   

In our experiments, we also consider computing the mean of the entire neighborhood.  Here we say $k=1$ to reinforce the connection to unary Janossy pooling
whereas with the LSTM model, $k$ refers to the number of samples of the neighborhood. 

In this section we investigate two conditions: (a) the impact of increasing $k$ in the $k$-ary dependencies; and (b) the benefits of increasing the number of sampled permutations at inference time.  To implement the model and design our experiments, we modified the reference PyTorch code provided by the authors\footnote{\footnotesize \url{https://github.com/williamleif/graphsage-simple/}, 
  see Appendix for details.}. 
 We consider the three graph datasets considered in~\citet{Hamilton2017}: Cora and Pubmed~\citep{sen2008collective}  and the larger Protein-Protein Interaction (PPI)~\citep{zitnik2017predicting}.  The first two are citation networks where vertices represent papers, edges represent citations, and vertex features are bag-of-words representations of the document text. The task is to classify the paper topic.  The PPI dataset is a collection of several graphs each representing human tissue; vertices represent proteins, edges represent protein interaction, features include genetic and immunological features, and we try to classify protein roles (there are 121 targets).  
More details of these experiments are shown in Table~\ref{tab:sumstats} in the \Appendix.    

(a) Table~\ref{tab:F1} shows the impact (on accuracy) of increasing the number of $k$-ary dependencies. 
We use $k_1, k_2 \in \{3, 5, 10, 25\}$ for the two pooling layers of our graph neural network (GNN).  The function $\harrow{f}$ is  an LSTM (except for when we try mean-pooling).  Note that for the LSTM, the number of parameters of the model is independent of $k$. At inference time, we sample 20 random permutations of each sequence and average the predicted probabilities before making a final prediction of the class label.    
The results in Table~\ref{tab:F1} show that the choice of $k_1,k_2\in \{3, 5, 10, 25 \}$  makes little difference on Cora and Pubmed due to the small neighborhood sizes: $k_1, k_2 \ge 5$ often amounts to sampling the entire neighborhood.  In PPI, whose average degree is 28.8, increasing $k$ yields consistent improvement.  The strong performance of mean-pooling points to both a relatively easy task\footnote{The topic of a paper can be adequately predicted by computing the average bag-of-words representations of papers in the neighborhood without reasoning about relationships between neighboring papers.} 
and the benefits of utilizing the entire neighborhood of each vertex.    
(b) We now investigate whether increasing the number of sampled permutations used to estimate \eqref{eq:yHat} at test (inference) time impacts accuracy. 
Figure~\ref{fig:performanceByPerm} in the \Appendix shows that increasing the number of sampled permutations from one to three leads to an increase in accuracy in the PPI task (Cora and Pubmed degrees are too small for this test) but diminishing returns set in by the seventh sample. Using paired tests -- t and Wilcoxon signed rank -- we see that test inference with seven sampled permutations versus one permutation is significant with $p < 10^{-3}$ over 12 replicates.  Sampling permutations at inference time is thus a cheap method for achieving modest but potentially important gains at inference time.
\begin{table}[t]
\vspace{-0pt}
\centering
\begin{minipage}[h]{.28\textwidth}
\caption{Accuracy (Micro-F1 score) using \Janossy pooling with $k$-ary dependencies and $\pi$-SGD in a graph neural network -- \graphsage \ -- with 20 permutations sampled at test time. Standard deviations over 30 runs for Cora/Pubmed and 4 runs for PPI are shown in parentheses.}
\label{tab:F1}
\end{minipage}
\vspace{-10pt}
\begin{minipage}[h]{.71\textwidth}
\begin{threeparttable}
\vspace{-15pt} 
\scalebox{0.9}{
\tabcolsep=0.11cm
\begin{tabular}{lrrrlll}
$\harrow{f}$ & method & $k_{1}$ & $k_{2}$  & \textbf{CORA} & \textbf{PUBMED} & \textbf{PPI}\tabularnewline
\hline 
LSTM & $\pi$-SGD &  3 & 3 & 0.860 (.009) & 0.889 (0.01) & 0.538 (.005)\tabularnewline
LSTM & $\pi$-SGD & 5 & 5 &   --\tnote{a}   & --\tnote{a}\tnote{a} & 0.579 (.015)\tabularnewline
LSTM & $\pi$-SGD & 10 & 25 &  --\tnote{a}    & --\tnote{a} & 0.650 (.013)\tabularnewline
LSTM & $\pi$-SGD & 25 & 10 &  --\tnote{a}    & --\tnote{a} & 0.689 (.062)\tabularnewline
LSTM & $\pi$-SGD & 25 & 25 &  --\tnote{a}    & --\tnote{a} & 0.702 (.044)\tabularnewline
LSTM & $\pi$-SGD & $|\vh|$ & $|\vh|$ &  --\tnote{a} & --\tnote{a} & 0.757 (.040) \tnote{b} \tabularnewline
Identity  & exact & 1 & 1 &  0.860 (.008) & 0.881 (.011) & 0.767 (.013)\tabularnewline
(mean-pool) & & & & & &
\end{tabular}
}
\begin{tablenotes}
\scriptsize
	\item [a] Entries denoted by -- all differ by less than 0.01. Typical neighborhoods in Cora and Pubmed are small, so that sampling $\ge 5$ neighbors is often equivalent to using the entire neighborhood.
	\item [b] Some neighbor sequences in PPI are prohibitively large, so we take $k_1 = k_2 = 100$.
\end{tablenotes}
\end{threeparttable}
\vspace{-10pt}
\end{minipage} 
\vspace{-10pt}
\end{table}
%
%
%
%
\section{Related Work}
\vspace{-5pt}
Under the \Janossy pooling framework presented in this work, existing literature falls under one of three approaches to approximating to the intractable 
\Janossy-pooling layer:
{\em Canonical orderings}, {\em $k$-ary dependencies}, and {\em permutation sampling}. We also discuss the broader context of invariant models and probabilistic interpretations.
\vspace{-5pt}
\paragraph{Canonical Ordering Approaches.} 
In \secref{sec:tract_canon}, we saw how permutation invariance can be achieved by mapping permutations to a canonical ordering. 
Rather than trying to define a good canonical ordering, one can try to learn it from the data, however
searching among all $|\vh|!$ permutations for one that correlates with the task of interest is a difficult discrete optimization problem. 
Recently, \citet{rezatofighi2018deep} proposed a method that computes the posterior distribution of all permutations, conditioned on the model and the 
data. This posterior-sampling approach is intractable for large inputs, unfortunately.
We note in passing that~\citet{rezatofighi2018deep} is interested in permutation-invariant outputs, and that \Janossy pooling is also trivially applicable to these tasks.
\citet{Vinyals2016} proposes a heuristic using ancestral sampling while learning the model.
\vspace{-5pt}
\paragraph{$k$-ary \Janossy Pooling Approaches.}
In \secref{sec:kAry} we described $k$-ary \Janossy pooling, which considers $k$-order relationships in the input vector $\vh$ to simplify optimization.
\deepsets~\citep{Zaheer2017} can be characterized as unary \Janossy pooling (i.e., $k$-ary for $k=1$). . \citet{qi2017pointnet} and
\citet{Ravanbakhsh2016} propose similar unary \Janossy pooling models.
\citet{cotter2018interpretable} proposes to add inductive biases to the {\em \deepsets} model in the form of monotonicity constraints with respect to the vector valued elements of the input sequence by modeling $f$ and $\rho$ with Deep Lattice Networks \citep{you2017deep}; one can extend~\citet{cotter2018interpretable} by using higher-order ($k > 1$) pooling. 

Exploiting dependencies within a sequence to learn a permutation-invariant function has been discussed elsewhere.  For instance~\citet{santoro2017simple} exploits pairwise relationships to perform relational reasoning about pairs of objects in an image and \citet{battaglia2018relational} contemplates modeling the center of mass of a solar system by including the pairwise interactions among planets.  However, \Janossy pooling provides a general framework for capturing dependencies within a permutation-invariant pooling layer.
\vspace{-5pt}
\paragraph{Permutation Sampling Approaches.}
In \secref{sec:PermSampling} we have seen a that permutation sampling can be used as a stochastic gradient procedure ($\pi$-SGD) to learn a model with a \Janossy pooling layer. The learned model provides only an approximate solution to original permutation-invariant function.
Permutation sampling has been used as a heuristic (without a theoretical justification) in both \citet{moore2017deep} and \citet{Hamilton2017}, which found that randomly permuting sequences and feeding them forward to an LSTM is effective in relational learning tasks that require permutation-invariant pooling layers. 
%
%
%
%
\vspace{-5pt}
\paragraph{Probabilistic Interpretation and Other Invariances}
Our work has a strong connection with finite exchangeability.
Some researchers may be more familiar with the 
concept of
{\em infinite exchangeability} through de Finetti's theorem~\citep{de1937prevision, diaconis1977finite}, which imposes strong structural requirements: the probability of any subsequence must equal the marginalized 
probability of the original sequence (projectivity).  \citet{korshunova2018bruno} noted the importance of this property for generative models and propose a model that learns a distribution without variational approximations.
Finite exchangeability drops this projectivity 
requirement~\citep{diaconis1977finite},
which in general, cannot be simplified beyond first sampling the number of observations $m$, and then sampling their locations from some exchangeable but 
non-i.i.d.\ distribution $p^m_{\text{exch}}(x_1,\ldots,x_m)$~\citep{daley2007introduction}. Equivalently, de Finetti's theorem for infinitely exchangeable sequences implies that the joint distribution can represented as a mixture distribution over conditionally independent random variables (given $\theta$)~\citep{de1937prevision, Orbanz2015} whereas the probability distribution of a finitely exchangeability sequence is a mixture over {\em dependent} random variables as shown by~\citet{diaconis1977finite}.  

In comparison, the restrictive assumption of letting $k=1$ in $k$-ary Janossy Pooling yields the form of a log-likelihood of conditionally iid random variables (consider $\harrow{f}$ a log pdf), the strong requirement of de Finetti's theorem for infinitely exchangeable sequences.  Conversely, higher-order Janossy pooling was designed to exploit dependencies among the random variables such as those that arise under finitely exchangeable distributions.  Indeed, finite exchangeability also arises from the theory of spatial point processes; our framework of Janossy pooling is inspired 
by \emph{Janossy densities}~\citep{daley2007introduction}, which model the finite exchangeable distributions as mixtures of non-exchangeable distributions applied to 
permutations. This literature also studies simplified exchangeable point processes such as finite Gibbs models \citep{vo2018model, moller2003statistical} 
that restrict the structure of $p_{\text{exch}}$ to fixed-order dependencies, and are related to $k$-ary \Janossy.

More broadly, there are other connections between permutation-invariant deterministic functions and exchangeability in probability distributions, as recently discussed by~\citet{bloem2019probabilistic}.  There, the authors also contemplate more general invariances through the language of group actions.  An example is {\em permutation equivariance}: one form of permutation equivariance asserts that ${f}(X_{\pi}) = {f}(X)_{\pi}\forall \pi \in \Pi_{|X|}$ where $f(X) $ is a sequence of length greater than 1. \citet{ravanbakhsh2017equivariance} provides a weight-sharing scheme for maintaining general neural network equivariances characterized as automorphisms of a colored multi-edged bipartite graph.  \citet{hartford2018deep} proposes a matrix completion model invariant to (possibly separate) permutations of the rows or columns.  Other invariances are studied through a probabilistic perspective in~\cite{Orbanz2015}.

\ifdefined\COMPLETE
\else
\documentclass[12pt]{article}
\usepackage{iclr2019_conference,times}
\newtheorem{thm}{Theorem}[section]

\usepackage{thmtools}
\usepackage{thm-restate}
\usepackage{hyperref}
\usepackage{wrapfig}

\usepackage[normalem]{ulem}
\usepackage{url}
\usepackage{multirow}
\usepackage{xspace}
\usepackage{amsmath}
\usepackage{cleveref}
\usepackage{amsthm}
\usepackage{textcomp}
\usepackage{amssymb}
\usepackage{amsfonts}
\usepackage[final]{graphicx}
\usepackage{threeparttable}
\usepackage{enumitem}
\usepackage{parskip}
\usepackage{mathrsfs}
\usepackage{booktabs}
\usepackage{caption}
\usepackage{tcolorbox}
\usepackage{times}
\usepackage{graphicx}
\usepackage{bbm}
\usepackage{color}
\usepackage{listings}

\usepackage{tikz}
\usepackage{tkz-graph}
\usetikzlibrary{positioning}
\usepackage{subfig}
\usepackage{booktabs}
\usepackage{makecell}
\usepackage{rotating}
\usepackage{pdflscape}
\usepackage{algorithm}
\usepackage[noend]{algpseudocode}
\usepackage{csquotes}
\usepackage{ulem}

\lstset{
basicstyle=\small\ttfamily,
columns=flexible,
breaklines=true
}

\begin{document}
\fi
\vspace{-5pt}
\section{Conclusions}
\vspace{-5pt}
Our approach of permutation-invariance through \Janossy pooling unifies a 
number of existing approaches, and opens up avenues to develop both new 
methodological extensions, as well as better theory. 
Our paper focused on two main approaches: $k$-ary interactions and 
random permutations. The former involves exact \Janossy pooling for a 
restricted class of functions $\harrow{f}$. Adding an additional neural 
network $\rho$ can recover lost model capacity and capture additional higher-order interactions, but hurts tractability and identifiability.  
Placing restrictions on $\rho$ (convexity, Lipschitz 
continuity etc.) can allow a more refined control of this trade-off, allowing 
theoretical and empirical work to shed light on the compromises involved. The second was a random permutation approach which conversely involves no clear trade-offs between model capacity and computation when $\rho$ is made more complex, instead it modifies the relationship between the tractable approximate loss $\dbar{J}$ and the original \Janossy loss $\dbar{L}$.
While there is a difference between $\dbar{J}$ and $\dbar{L}$, we saw the strongest empirical performance coming from this approach in our experiments (shown in the last row of Table~\ref{tab:accuracy}); future work is required to identify which problems $\pi$-SGD is best suited for and when its convergence criteria are satisfied.
Further, a better understanding how the loss-functions $\dbar{L}$ and $\dbar{J}$ relate to each other can shed light 
on the slightly black-box nature of this procedure. It is also important to understand the relationship between
the random permutation optimization to canonical ordering and how one might be used to improve the other. 
Finally, it is important to apply our methodology to a wider range of applications. Two immediate domains are more challenging tasks involving 
graphs and tasks involving non-Poisson point processes.

\ifdefined\COMPLETE
\else
\end{document}
\fi

\subsubsection*{Acknowledgments}
This work was sponsored in part by the ARO, under the U.S. Army Research Laboratory contract number W911NF-09-2-0053, by the Purdue Integrative Data Science Initiative and Purdue Research foundation, the DOD through SERC under Contract No. HQ0034-13-D-0004 RT \#206, the National Science Foundation under contract numbers IIS-1816499 and DMS-1812197, and the NVIDIA GPU grant program for hardware donation.  


\bibliography{Graphs2018,iclr_rlm19,AH_Ribeiro,ryans_bib}
\bibliographystyle{iclr2019_conference}

%
\renewcommand\thesection{\Alph{section}}
\setcounter{section}{0}

\section{Proofs of Results}

We restate and prove Proposition~\ref{prop:kPerms}.

\karytract*
\begin{proof}
	Define two permutations $\pi, \pi' \in \Pi_{|\vhid|}$ that agree on the 
	first $k$ elements as {\em $k$-equivalent}. Such permutations satisfy 
	$\harrow{f}(\kcard, \pj{k}(\vhid_{\pi}); \ftheta) = \harrow{f}(\kcard, \pj{k}(\vhid_{\pi'}); \ftheta)$. 
	These two permutations belong to the same equivalence class, containing 
	a total of $(|\vhid|-k)!$ permutations (obtained by permuting the last $(|\vhid|-k)$ 
	elements). Overall, we then have a total of $|\vhid|!/(|\vhid|-k)!$ equivalence 
	classes. Write the set of equivalence classes as $\Pi_{|\vhid|}^k$, and 
	represent each by one of its elements. Then,
	\begin{align}
		\label{eq:FastK} 
		\dbar{f}(|\vhid|, \vhid ; \ftheta ) &= \frac{1}{|\vhid|!} \sum_{\pi \in \Pi_{|\vhid|}} \harrow{f}(\kcard, \pj{k}(\vhid_{\pi}); \ftheta) =\frac{(|\vhid|-k)!}{|\vhid|!}\sum_{\pi\in \Pi_{|\vhid|}^k}\harrow{f}\big(\kcard,\pj{k}(\vhid_\pi) ; \ftheta \big)  \nonumber
	\end{align}
	is now a summation over only $|\vhid|!/(|\vhid|-k)!$ terms.  We can conclude that
	\begin{equation*}
		\dbar{f}(|\vhid|, \vhid ; \ftheta )= \frac{(|\vhid|-k)!}{|\vhid|!}\sum_{(i_1, i_2, \ldots, i_k) \in \sI_{|\vhid|}}\harrow{f}\big(\kcard, (\evhid_{i_1}, \evhid_{i_2}, \ldots, \evhid_{i_k}) ; \ftheta \big).%
	\end{equation*}
\end{proof}%
Next, we restate and prove the remaining portion of Theorem~\ref{thm:k_implies_km}.
\staircase*
\begin{proof}~\\
($\mathcal{F}_{k-1}\subset\mathcal{F}_{k}$): %
Consider any element 
$\dbar{f}_{k-1} \in \mathcal{F}_{k-1}$, and write $\harrow{f}(\kcard,~\cdot~; \btheta^{(f)})$ for its 
associated Janossy function.
For any sequence $\vhid$, %
$\harrow{f}(\kcard, \pj{k-1}(\vhid); \ftheta) = 
\harrow{f}(\kcard, \pj{k-1}(\pj{k}(\vhid)); \ftheta) := 
\harrow{f}_{+}(\kcard, \pj{k} (\vhid); \ftheta)$, where the function 
$\harrow{f}_+$ 
looks at its first $k$ elements.
Thus, 
\begin{align}
	\dbar{f}_{k-1}(|\vhid|, \vhid ; \ftheta ) &
	= \frac{1}{|\vhid|!} \sum_{\pi \in \Pi_{|\vhid|}} \harrow{f}(\kcard, {\pj{k-1\:}}(\vhid_{\pi}); \ftheta)   
	= \frac{1}{|\vhid|!} \sum_{\pi \in \Pi_{|\vhid|}} \harrow{f}_{+}(\kcard, \pj{k}(\vhid_{\pi}); \ftheta) \nonumber \\ 
	&= \dbar{f}_{k}(|\vhid|, \vhid ; \ftheta ),
	\vspace{-10pt}
\end{align}
where $\dbar{f}_{k}$ is the \Janossy function associated with $\harrow{f}_{+}$ and thus belongs to 
$\mathcal{F}_k$.	

%
%
($\mathcal{F}_{k}\not\subset\mathcal{F}_{k-1}$): %
the case where $k=1$ is trivial, so assume $k > 1$.  We will demonstrate the existence of $\dbar{f}_{k}\in\mathcal{F}_{k}$
such that $\dbar{f}_{k-1}\ne\dbar{f}_{k}$ for all
$\dbar{f}_{k-1}\in\mathcal{F}_{k-1}$.  Let $\dbar{f}_{k}$ and $\dbar{f}_{k-1}$ be associated with $\harrow{f}_{k}$ and $\harrow{f}_{k-1}$, respectively.  

It suffices to consider
$|\vhid|=k$.  Let $\harrow f_{k}(|\vhid|,\vhid_{\pi} ; \thetaToF_{k})=\prod_{l=1}^{|\vhid|}\evhid_{\pi(l)}$
whence $\dbar{f}_{k}(|\vhid|, \vhid ; \thetaToF_{k})=\prod_{l=1}^{|\vhid|}\evhid_{l}$.
Thus, for any $\dbar{f}_{k-1}$ and any $\thetaToF_{k-1}$,
\begin{align*}
\dfrac{\dbar{f}_{k-1}(|\vhid|,\vhid ; \thetaToF_{k-1})}%
{\dbar{f}_{k}(|\vhid|,\vhid; \thetaToF_k)}
 & =\frac{1}{|\vhid|!}\sum_{\pi\in \Pi_{|\vhid|}}\dfrac{\harrow{f}_{k-1}(\kcard, \pj{k-1}(\vhid_{\pi}); \thetaToF_{k-1})}{\prod_{l=1}^{|\vhid|}\evhid_{l}}\\ %
 & =\frac{1}{|\vhid|!}\sum_{j=1}^{|\vhid|}\sum_{\tilde{\pi}\in \Pi_{\{1,\ldots,|\vhid|\}\setminus{j}}}\dfrac{\harrow{f}_{k-1}\big(\kcard,(\vhid_{-j})_{\tilde{\pi}} ; \thetaToF_{k-1} \big)}{\prod_{l=1}^{|\vhid|}\evhid_{l}}
\end{align*}
where $\Pi_{\{1,\ldots,|\vhid|\}\setminus{j}}$ denotes the set of permutation functions defined on $\{ 1, 2, \ldots, j-1, j+1, \ldots, |\vhid| \}$ and $(\vhid_{-j})_{\tilde{\pi}}$ is a permutation of the sequence $\big(\evhid_{1}, \ldots, \evhid_{j-1}, \evhid_{j+1},\ldots, \evhid_{|\vhid|}\big)$.  %
This can be written as
\[
\frac{1}{|\vhid|!}\sum_{j=1}^{|\vhid|}\frac{1}{\evhid_{j}}\underbrace{\Big(\sum_{\tilde{\pi}\in \Pi_{\{1,\ldots,|\vhid|\}\setminus{j}}}\dfrac{\harrow{f}_{k-1}\big(\kcard,(\vhid_{-j})_{\tilde{\pi}} ; \thetaToF_{k-1} \big)}{\prod_{l\ne j}\evhid_{l}}\Big)}_{\text{denote by } a_{j,|\vhid|}},
\]
therefore
\begin{equation}
\dfrac{\dbar{f}_{k-1}(|\vhid|,\vhid_{\pi} ; \thetaToF_{k-1})}{\dbar{f}_{k}(|\vhid|,\vhid_{\pi}; \thetaToF_k)}=\frac{1}{|\vhid|!}\sum_{j=1}^{|\vhid|}\frac{1}{\evhid_{j}}a_{j,|\vhid|}.\label{eq:KeyQuotient}
\end{equation}
Now, $\dbar{f}_{k-1}=\dbar{f}_{k}$ if and only
if their quotient in~\eqref{eq:KeyQuotient} is unity for all $\vhid$.
But this is clearly not possible in general unless $\sH$
is a singleton, which we have precluded in our assumptions.
\end{proof}

%
%
%
%
%

Proposition~\ref{p:piSGD} is repeated below and is followed by a more rigorous restatement.

\pisgd*

The following statement is similar to that in~\citet{Yuille2004}, which also provides intuition behind the theoretical assumptions, which are indeed quite general.  See also~\citep{younes1999convergence}.  This is a familiar application of stochastic approximation algorithms already used in training neural networks.

\begin{proposition}[$\pi$-SGD Convergence]\label{p:piSGDFormal}
	Consider the $\pi$-SGD algorithm in Definition~\ref{d:piSGD}.
	If
	\begin{enumerate}[leftmargin=.25in]
		\item[(a)] there exists a constant $M > 0$ such that for all $\vtheta$, $- \mG_t^T \vtheta \leq M \Vert \vtheta - \vtheta^\star \Vert_2^2$,
		where $\mG_t$ is the true gradient for the full batch over all permutations,
		$
		\mG_t = \nabla_\vtheta \dbar{\Jloss}(\train;\vtheta^{(\rho)}_t, \vtheta^{(f)}_t, \vtheta^{(h)}_t),
		$
		where $\vtheta \equiv (\vtheta^{(\rho)}, \vtheta^{(f)}, \vtheta^{(h)})$, and $\vtheta^\star$ is the optimum.
		
		\item[(b)] there exists a constant $\delta > 0$ such that for all $\vtheta$, $\expected_t [ \Vert \rmZ_t\Vert_2^2] \leq \delta^2 (1 +\Vert \vtheta_{t} - \vtheta^{\star}_{t} \Vert_2^2)$, where the expectation is taken with respect to all the data prior to step $t$.
	\end{enumerate}
	Then, the algorithm in \eqref{eq:piSGD} converges to $\vtheta^\star$ with probability one.
\end{proposition}
\begin{proof}
	
	First, we can show that $\expected_t[\rmZ_t] = \mG_t$ by \eqref{eq:RLoss}, the linearity of the derivative operator, and the fact that the permutations are independently sampled for each training example in the mini-batch and are assumed independent of $\vtheta$.
	That \eqref{eq:piSGD} converges to $\vtheta^\star$ is a consequence of our conditions and the supermartingale convergence theorem~\citep[pp.\ 481]{grimmett2001probability}.
	The following argument follows \citet{Yuille2004}.
	Let $A_t =  \Vert \vtheta_t - \vtheta^\star \Vert_2^2$, $B_t =  \delta^2 \eta_t^2$, and $C_t = -\Vert \vtheta_t - \vtheta^\star \Vert_2^2 (\delta^2 \eta_t^2 - 2 M \eta_t)$.  Note that $C_t$ is positive for a sufficiently large $t$, and $\sum_{t=1}^\infty B_t \leq \infty$ by our definition of $\eta_t$ (Definition ~\ref{d:piSGD}).
	We will demonstrate that $\expected_{t}[A_t] \leq A_{t-1} + B_{t-1} - C_{t-1}$, for all $t$, in the \Appendix from which it follows that $A_t$ converges to zero with probability one and $\sum_{t=1}^\infty C_t < \infty$.
	We write
	\begin{align*}
	\expected_{t} \left[\Vert  \vtheta_{t}  - \vtheta^\star \Vert_2^2 \right] 
	&= \expected_{t} \left[\Vert  \vtheta_{t-1} - \eta_{t-1} \rmZ_{t-1}  - \vtheta^\star \Vert_2^2 \right] \\
	&= \Vert  \vtheta_{t-1} - \vtheta^\star \Vert_2^2  - 2 \eta_{t-1} \expected_{t}[ (\vtheta_{t-1} - \vtheta^\star)^T \rmZ_{t-1}] + \eta_{t-1}^2 \expected_{t} [\Vert \rmZ_{t-1} \Vert_2^2 ]  \\
	&\leq \Vert  \vtheta_{t-1} - \vtheta^\star \Vert_2^2  - 2 \eta_{t-1} (\vtheta_{t-1} - \vtheta^\star)^{T} \mG_{t-1}+ \delta^{2}\eta_{t-1}^2  +  \delta^{2} \eta_{t-1}^2\Vert  \vtheta_{t-1} - \vtheta^\star \Vert_2^2 \\
	&\leq \Vert  \vtheta_{t-1} - \vtheta^\star \Vert_2^2  - 2 M \eta_{t-1}  \Vert \vtheta_{t-1} - \vtheta^\star\Vert_2^2 +  \delta^{2}\eta_{t-1}^2 +  \delta^{2} \eta_{t-1}^2 \Vert  \vtheta_{t-1} - \vtheta^\star \Vert_2^2 ,		
	\end{align*}
	and the result follows.
\end{proof}


\section{Experiments: Further Results and Implementation Details}
\subsection{Results}
\label{sup:Results}
The accuracy scores for all models (including the LSTM) on the sequence arithmetic tasks are shown in Table~\ref{tab:accuracyFull}.  This table repeats results shown in Table~\ref{tab:accuracy}, except here we show additional rows representing models that use LSTM as $\harrow{f}$.  We chose accuracy (0-1 loss) to be consistent with~\citet{Zaheer2017}; here we report mean absolute error to evaluate the differences it makes on our results.  These can be found in Tables~\ref{tab:mae} and~\ref{tab:mae_variance}.  The message is similar to the one told by accuracy scores; there is a drop in the mean absolute error as the value of $k$ increases and when using more sampled permutations at test-time (e.g., \Janossy-20inf-LSTM versus \Janossy-1inf-LSTM).  Again, the power of using an RNN for $\harrow{f}$ and training with $\pi$-SGD is salient on the variance task where it is important to exploit dependencies in the sequence.  Beyond the performance gains, we also observe a drop in variance when sampling more permutations at test time.  Furthermore, as discussed in the implementation section, we constructed $k$-ary models to have the same number of parameters regardless of $k$ for the results reported in the main body.  We show results where this constraint is relaxed in Table~\ref{tab:inc_accuracy}. Here we see a modest improvement of $k$-ary models which stands to reason considering the embedding dimension fed to the Janossy pooling layer was reduced from 100 with $k=1$ to 33 with $k=3$ (please see the implementation section for details).  

\begin{figure}[t]
	\vspace{-5pt}
	\centering 
	\includegraphics[width=0.60\textwidth]{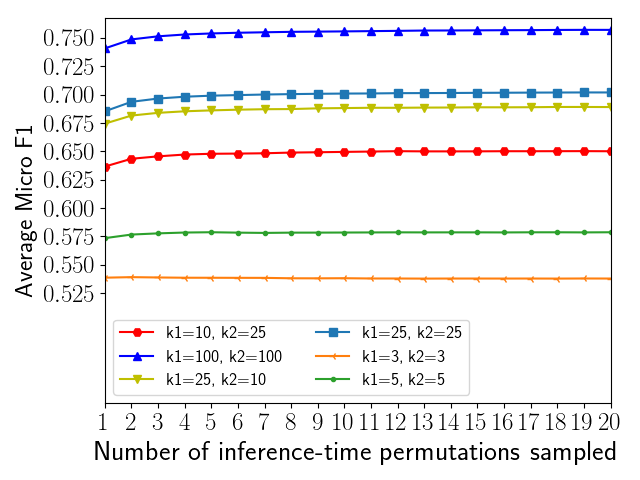}
	\vspace{-15pt}
	\caption{Mean performance vs number of permutations sampled at test time, PPI task 	\vspace{-15pt}} 
	\label{fig:performanceByPerm} 
\end{figure}

For the graph tasks, the plot of performance as a function of number of inference-time permutations is shown in Figure~\ref{fig:performanceByPerm}.%
{
\begin{table}[b!!]
  \small
\vspace{-5pt}
\centering
\caption{Full table showing the Accuracy (and RMSE for the {\em variance} task) for all models used for the sequence arithmetic tasks. The {\em method} column refers to the method used to deal with the sum over all permutations. {\em Infr sample} refers to the number of permutations sampled at test time to estimate \eqref{eq:yHat} for methods learned with $\pi$-SGD. $k=1$ corresponds to \deepsets. $\tanh$ activations are used with the MLP. Standard deviations computed over 15 runs are shown in parentheses.
}
\label{tab:accuracyFull}
\vspace{-5pt}
\scalebox{0.98}{
\tabcolsep=0.11cm
\begin{tabular}{llrclllllr}
\multicolumn{1}{c}{\em $\harrow{f}$} &\multicolumn{1}{p{1cm}}{\em \centering method} &\multicolumn{1}{p{0.8cm}}{\em \centering infr \\ sample} &\multicolumn{1}{c}{\em k} &\multicolumn{1}{c}{\em $\rho$} &\multicolumn{1}{c}{\em sum} &\multicolumn{1}{c}{\em range} &{\em \centering unique sum} &{\em \centering uniq.\ count}&{\em \centering variance}
\\ \hline
MLP (30) & exact & \NA & \centering{1} & Linear          & 1.00(0.00)     & 0.04(0.00) & 0.07(0.00) & 0.36(0.01) & 119.05(1.29) \\
MLP (30) & exact & \NA & \centering{2} & Linear          & 0.99(0.00) & 0.09(0.00) & 0.17(0.00) & 0.74(0.03)     & 4.37(0.50)     \\
MLP (30) & exact & \NA & \centering{3} & Linear          & 0.99(0.00)  & 0.21(0.00) & 0.44(0.02) & 0.89(0.04)  & 8.99(0.99)    \\
MLP (30) & exact & \NA & \centering{1} & MLP (100) & 1.00(0.00)   & 0.97(0.01) & 1.00(0.00)     & 1.00(0.00)     & 1.95(0.24)      \\
MLP (30) & exact & \NA & \centering{2} & MLP (100) & 1.00(0.00) & 0.97(0.01) & 1.00(0.00)     & 1.00(0.00)     & 3.49(0.48)  \\
MLP (30) & exact & \NA & \centering{3} & MLP (100) & 0.93(0.02) & 0.93(0.02) & 1.00(0.00)     & 1.00(0.00)     & 6.90(0.47) \\
LSTM(50) & $\pi$-SGD & 1 & \centering{$|\vh|$} & Linear          & 0.99(0.00) & 0.95(0.01) & 1.00(0.00)     & 1.00(0.00)     & 1.65(0.22)  \\    
LSTM(50) & $\pi$-SGD & 20 & \centering{$|\vh|$} & Linear          & 0.99(0.00)   & 0.97(0.01)  & 1.00(0.00)     & 1.00(0.00)     & 1.39(0.26)   \\   
GRU(80) & $\pi$-SGD & 1 & \centering{$|\vh|$} & Linear          & 0.99(0.01) & 0.98(0.00) & 1.00(0.00)     & 1.00(0.00)     & 1.43(0.23)      \\
GRU(80) & $\pi$-SGD & 20 & \centering{$|\vh|$} & Linear          & 0.99(0.00) & 0.99(0.00) & 1.00(0.00)     & 1.00(0.00)     & 1.20(0.23)   \\   
LSTM(50) & $\pi$-SGD & 1 & \centering{$|\vh|$} & MLP (100) & 0.99(0.01)  & 0.99(0.00) & 1.00(0.00)     & 1.00(0.00)     & 1.05(0.77)      \\
LSTM(50) & $\pi$-SGD & 20 & \centering{$|\vh|$} & MLP (100) & 0.99(0.00) & 1.00(0.00) & 1.00(0.00)     & 1.00(0.00)     & 1.02(0.41)    \\  
GRU(80) & $\pi$-SGD & 1 & \centering{$|\vh|$} & MLP (100) & 0.99(0.00) & 1.00(0.00) & 1.00(0.00)     & 1.00(0.00)     & 0.42(0.62)    \\
GRU(80) & $\pi$-SGD & 20 & \centering{$|\vh|$} & MLP (100) & 0.99(0.00) & 1.00(0.00) & 1.00(0.00)     & 1.00(0.00)     & 0.40(0.37)     
\end{tabular}
}
\vspace{-10pt}
\end{table}
}

\begin{table}[b!!]
\noindent
\caption{Mean Absolute Error of various \Janossy pooling approximations under distinct tasks. The column method refers to the tractability strategy. Inf sample refers to the number of permutations sampled to estimate \eqref{eq:yHat} for methods learned with $\pi$-SGD. $k=1$ corresponds to \deepsets. $\tanh$ activations are used with the MLP's. Standard deviations computed over 15 runs are shown in parentheses.
}
\label{tab:mae}
\scalebox{0.87}{
\begin{tabular}{llrclrrrr}
\multicolumn{1}{c}{\em $\harrow{f}$} &\multicolumn{1}{p{1cm}}{\em \centering method} &\multicolumn{1}{p{1cm}}{\em \centering inf \\ sample} &\multicolumn{1}{c}{\em k} &\multicolumn{1}{c}{\em $\rho$} &\multicolumn{1}{c}{\em sum} &\multicolumn{1}{c}{\em range} &{\em \centering unique sum} &{\em \centering unique count}
\\ \hline
MLP (30) & exact & -- & 1 & Linear          & 0.000(0.000)     & 9.366(0.094) & 4.209(0.025) & 0.828(0.008)  \\
MLP (30) & exact & -- & 2 & Linear          & 0.006(0.011) & 4.143(0.041) & 1.968(0.016) & 0.277(0.029)  \\
MLP (30) & exact & -- & 3 & Linear          & 0.037(0.031) & 2.307(0.074) & 0.730(0.040)   & 0.114(0.040)    \\
MLP (30) & exact & -- & 1 & MLP (100) & 0.001(0.000) & 0.033(0.003)  & 0.000(0.000)     & 0.000(0.000)      \\
MLP (30) & exact & -- & 2 & MLP (100) & 0.007(0.005) & 0.038(0.006) & 0.000(0.000)     & 0.000(0.000)       \\
MLP (30) & exact & -- & 3 & MLP (100) & 0.091(0.026)   & 0.147(0.049) & 0.000(0.000)     & 0.000(0.000)     \\
LSTM(50) & $\pi$-SGD & 1 & $|\vh|$  & Linear          & 0.003(0.002) & 0.051(0.010)  & 0.000(0.000)    & 0.000(0.000)        \\
LSTM(50) & $\pi$-SGD & 20 & $|\vh|$  & Linear          & 0.001(0.001) & 0.035(0.006) & 0.000(0.000)    & 0.000(0.000)      \\
GRU(80) & $\pi$-SGD & 1 & $|\vh|$  & Linear          & 0.007(0.012) & 0.020(0.005)  & 0.000(0.000)    & 0.000(0.000)    \\
GRU(80) & $\pi$-SGD & 20 & $|\vh|$  & Linear          & 0.001(0.002) & 0.014(0.004) & 0.000(0.000)     & 0.000(0.000)     \\
LSTM(50) & $\pi$-SGD & 1 & $|\vh|$  & MLP (100) & 0.007(0.010)  & 0.006(0.001) & 0.000(0.000)    & 0.000(0.000)      \\
LSTM(50) & $\pi$-SGD & 20 & $|\vh|$  & MLP (100) & 0.004(0.006) & 0.005(0.001) & 0.000(0.000)     & 0.000(0.000)    \\
GRU(80) & $\pi$-SGD & 1  & $|\vh|$  & MLP (100) & 0.002(0.004) & 0.002(0.001) & 0.000(0.000)     & 0.000(0.000)      \\
GRU(80) & $\pi$-SGD & 20  & $|\vh|$ & MLP (100) & 0.002(0.003) & 0.002(0.001) & 0.000(0.000)     & 0.000(0.000)    \\
\end{tabular}
}
\end{table}

\begin{table}[t]
\noindent
\caption{Mean Absolute Error of various \Janossy pooling approximations for the variance task. The column method refers to the tractability strategy. Inf sample refers to the number of permutations sampled to estimate \eqref{eq:yHat} for methods learned with $\pi$-SGD. $k=1$ corresponds to \deepsets. $\tanh$ activations are used with the MLP's. Standard deviations computed over 15 runs are shown in parentheses.
}
\label{tab:mae_variance}
\begin{center}
\begin{tabular}{llrclr}
\multicolumn{1}{c}{\em $\harrow{f}$} &\multicolumn{1}{p{1cm}}{\em \centering method} &{\em  inf  sample} &\multicolumn{1}{c}{\em k} &\multicolumn{1}{c}{\em $\rho$}  &{\em \centering variance}
\\ \hline
MLP (30) & exact & --& 1 & Linear          & 69.953(0.492) \\
MLP (30) & exact & -- & 2 & Linear          & 2.262(0.363)  \\
MLP (30) & exact & -- & 3 & Linear          & 6.747(0.871)  \\
MLP (30) & exact & -- & 1 & MLP (100) & 0.613(0.107)  \\
MLP (30) & exact & -- & 2 & MLP (100) & 1.733(0.146)  \\
MLP (30) & exact & -- & 3 & MLP (100) & 4.379(0.318)  \\
LSTM(50) & $\pi$-SGD & 1 & $|\vh|$  & Linear        & 0.801(0.200)    \\
LSTM(50) & $\pi$-SGD & 20 & $|\vh|$ & Linear          & 0.698(0.412)  \\
GRU(80) & $\pi$-SGD & 1 & $|\vh|$ & Linear          & 0.795(0.205)  \\
GRU(80) & $\pi$-SGD & 20 & $|\vh|$ & Linear          & 0.672(0.332)  \\
LSTM(50) & $\pi$-SGD & 1 & $|\vh|$ & MLP (100) & 0.604(0.078)  \\
LSTM(50) & $\pi$-SGD & 20 & $|\vh|$ & MLP (100) & 0.422(0.102)  \\
GRU(80) & $\pi$-SGD & 1  & $|\vh|$ & MLP (100) & 0.594(0.634)  \\
GRU(80) & $\pi$-SGD & 20  & $|\vh|$ & MLP (100) & 0.517(0.084) \\
\end{tabular}
\end{center}
\end{table}

\begin{table}[t]
  \small
\vspace{-5pt}
\centering
\caption{Accuracy (and RMSE for the {\em variance} task) of $k-ary$ \Janossy pooling approximations with the same input dimension as $k=1$ under distinct tasks. The {\em method} column refers to the method used to deal with the sum over all permutations. {\em Infr sample} refers to the number of permutations sampled at test time to estimate \eqref{eq:yHat} for methods learned with $\pi$-SGD. $k=1$ corresponds to \deepsets. $\tanh$ activations are used with the MLP. Standard deviations computed over 15 runs are shown in parentheses.}
\label{tab:inc_accuracy}
\scalebox{0.96}{
\tabcolsep=0.11cm
\begin{tabular}{llllllllr}	
	\multicolumn{1}{c}{\em $\harrow{f}$} &\multicolumn{1}{p{1cm}}{\em \centering method} &\multicolumn{1}{c}{\em k} &\multicolumn{1}{c}{\em $\rho$} &\multicolumn{1}{c}{\em sum} &\multicolumn{1}{c}{\em range} &{\em \centering unique sum} &{\em \centering uniq.\ count}&\multicolumn{1}{c}{\em \centering variance}
	\\ \hline
	MLP (30) & exact  & \centering{1~~} & Linear          & 1.00(0.00)     & 0.04(0.00) & 0.07(0.00) & 0.36(0.01) & 119.05(1.29) \\
	MLP (30) & exact & \centering{2$*$} & Linear          & 1.00(0.00) & 0.09(0.00) & 0.18(0.00) & 0.74(0.03)     & 0.71(0.04)     \\
	MLP (30) & exact & \centering{3$*$} & Linear          & 1.00(0.00)  & 0.22(0.00) & 0.51(0.01) & 0.98(0.00)  & 1.54(0.99)    \\
	MLP (30) & exact & \centering{1~~} & MLP (100) & 1.00(0.00)   & 0.97(0.01) & 1.00(0.00)     & 1.00(0.00)     & 1.95(0.24)      \\
	MLP (30) & exact & \centering{2$*$} & MLP (100) & 1.00(0.00) & 0.99(0.00) & 1.00(0.00)     & 1.00(0.00)     & 2.65(0.50)  \\
	MLP (30) & exact & \centering{3$*$} & MLP (100) & 1.00(0.00) & 0.99(0.00) & 1.00(0.00)     & 1.00(0.00)     & 3.44(0.51) \\
\end{tabular}
}
\end{table}

\begin{table}[t]
  \small
\vspace{-5pt}
\centering
\caption{Mean Absolute Error of $k-ary$ \Janossy pooling approximations with the same input dimension as $k=1$ under distinct tasks. The {\em method} column refers to the method used to deal with the sum over all permutations. {\em Infr sample} refers to the number of permutations sampled at test time to estimate \eqref{eq:yHat} for methods learned with $\pi$-SGD. $\tanh$ activations are used with the MLP. Standard deviations computed over 15 runs are shown in parentheses.
}
\label{tab:inc_mae}
\scalebox{0.96}{
\tabcolsep=0.11cm
\begin{tabular}{llllllllr}
\multicolumn{1}{c}{\em $\harrow{f}$} &\multicolumn{1}{p{1cm}}{\em \centering method}  &\multicolumn{1}{c}{\em k} &\multicolumn{1}{c}{\em $\rho$} &\multicolumn{1}{c}{\em sum} &\multicolumn{1}{c}{\em range} &{\em \centering unique sum} &{\em \centering uniq.\ count}&{\em \centering variance}
\\ \hline
MLP (30) & exact & \centering{1} & Linear          & 0.00(0.00)     & 9.37(0.09) & 4.21(0.03) & 0.83(0.01) & 69.95(0.49) \\
MLP (30) & exact & \centering{$2^*$} & Linear          & 0.00(0.00) & 4.12(0.05) & 1.95(0.01) & 0.29(0.03)     & 0.46(0.04)     \\
MLP (30) & exact  & \centering{$3^*$} & Linear          & 0.00(0.00)  & 2.31(0.04) & 0.64(0.02) & 0.02(0.00)  & 1.09(0.11)    \\
MLP (30) & exact  & \centering{1} & Linear          & 0.00(0.00)     & 0.03(0.00) & 0.00(0.00) & 0.00(0.00) & 0.61(0.10) \\
MLP (30) & exact  & \centering{$2^*$} & MLP (100) & 0.00(0.00) & 0.01(0.00) & 0.00(0.00)     & 0.00(0.00)     & 0.96(0.09)  \\
MLP (30) & exact  & \centering{$3^*$} & MLP (100) & 0.02(0.00) & 0.02(0.00) & 0.00(0.00)     & 0.00(0.00)     & 1.39(0.12) \\
\end{tabular}
}
\end{table}
\clearpage

\subsection{Implementation and Experiment Details}
\label{sup:ImplDetails}

\paragraph{Sequence tasks}

We extended the code from~\citet{Zaheer2017}, which was written in Keras\citep{chollet2015keras}, and subsequently ported to PyTorch.  For $k$-ary models with $k \in \{2, 3\}$, we always sort the sequence $\vdata$ beforehand to reduce the number of combinations we need to sum over.  In the notation of Figure~\ref{fig:architecture}, $\vh$ is an Embedding with dimension of $\mathrm{floor}(\frac{100}{k}) $ (to keep the total number of parameters consistent for each $k$ as discussed below), $\harrow{f}$ is either an MLP with a single hidden layer or an RNN depending on the model ($k$-ary \Janossy or full-\Janossy, respectively), and $\rho$ is either a linear dense layer or one hidden layer followed by a linear dense layer.  The MLPs in $\harrow{f}$ have 30 neurons whereas the MLPs in $\rho$ have 100 neurons, the LSTMs have 50 neurons, and the GRUs have 80 hidden neurons.  All activations are $\mathrm{tanh}$ except for the output layer which is linear.   We chose 100 for the embedding dimension to be consistent with~\citet{Zaheer2017}.

 For the $k$-ary results shown in the body, we made sure the number of parameters was consistent for $k \in \{1, 2, 3 \}$ (see Table~\ref{tab:num_parameters}).  We unify the number of parameters by adjusting the output dimension of the embedding.  We also experimented with relaxing the restriction that $k$-ary models have the same numbers of parameters (Table~\ref{tab:inc_accuracy}), and the numbers of parameters in these models is also shown in Table~\ref{tab:num_parameters}.  For the LSTM than GRU models, we follow the choice of~\citet{Zaheer2017} which also reports that the choices were made to keep the numbers of parameters consistent.

Optimization is done with Adam~\citep{KingmaAdam} with a tuned the learning rate, searching over $\{0.01,0.001,0.0001,0.00001\}$.  Training was performed on GeForce GTX 1080 Ti GPUs.  

\vspace{-5pt}
\paragraph{Graph-based tasks}
The datasets used for this task are summarized in Table~\ref{tab:sumstats}.
Our implementation is in PyTorch using Python 2.7, following the PyTorch code associated with~\citet{Hamilton2017}.  That repo did not include an LSTM aggregator, so we implemented our own following the TensorFlow implementation of \graphsage, and describe it here.  At the beginning of every forward pass, each vertex $v$ is associated with a $p$-dimensional vertex attribute $\vhid$ (see~Table\ref{tab:sumstats}).  For every vertex in a batch, $k_1$ neighbors of $v$ are sampled, their order is shuffled, and their features are fed through an LSTM.  From the LSTM, we take the short-term hidden state associated with the last element in the input sequence (denoted $\vh_{(T)}$ in the LSTM literature, but this $\vh$ is not to be confused with a vertex attribute).  This short-term hidden state is passed through a fully connected layer to yield a vector of dimension $\frac{q}{2}$, where $q$ is a user-specified positive even integer referred to as the \emph{embedding dimension}.  The vertex's own attribute $\vhid$ is also fed forward through a fully connected layer with $\frac{q}{2}$ output neurons.  At this point, for each vertex, we have two representation vectors of size $\frac{q}{2}$ representing the vertex $v$ and its neighbor multiset, which we concatenate to form an embedding of size $q$.  This describes one convolution layer, and it is repeated a second time with a distinct set of learnable weights for the fully connected and LSTM layers, sampling $k_2$ vertices from each neighborhood and using the embeddings of the first layer as features.  After each convolution, we may optionally apply a ReLU activation and/or embedding normalization, and we follow the decisions shown in the \graphsage code ~\citet{Hamilton2017}.  After both convolution operations, we apply a final fully connected layer to obtain the score, followed by a softmax (Cora, Pubmed) or sigmoid (PPI).  The loss function is cross entropy for Cora and Pubmed, and binary cross entropy for PPI.  

\clearpage

\begin{table}[t]
	\noindent
	\caption{Number of trainable parameters in each of the $k$-ary Janossy Pooling approaches. The $k$-ary models indicated with a $*$ take 100 dimensional embeddings as input to $\harrowStable{f}$, in contrast with the approach taken in Table~\ref{tab:accuracy} where the embedding was of size $\mathrm{floor}(100/k)$. }
	\label{tab:num_parameters}
	\begin{center}
		\begin{tabular}{lllr}
			\multicolumn{1}{c}{\em $\harrow{f}$}  &\multicolumn{1}{c}{\em k} &\multicolumn{1}{c}{\em $\rho$}  &\multicolumn{1}{p{2.5cm}}{\em \centering \#  trainable parameters}
			\\ \hline
			MLP (30) & 1  & Linear          & 3061                                                                     \\
			MLP (30) & 2  & Linear          & 3061                                                                     \\
			MLP (30) & 3  & Linear          & 3031                                                                     \\
			MLP (30) & $2^*$ & Linear          & 6061                                                                     \\
			MLP (30) & $3^*$ & Linear          & 9061                                                                     \\
			MLP (30) & 1  & MLP (100) & 6231                                                                     \\
			MLP (30) & 2  & MLP (100) & 6231                                                                     \\
			MLP (30) & 3  & MLP (100) & 6201                                                                     \\
			MLP (30) & $2^*$ & MLP (100) & 9231                                                                     \\
			MLP (30) & $3^*$ & MLP (100) & 12231                                                                    \\
			LSTM(50)      & n  & Linear          & 30451                                                                    \\
			GRU(80)       & n  & Linear          & 43761                                                                    \\
			LSTM(50)      & n  & MLP (100) & 35601                                                                    \\
			GRU(80)       & n  & MLP (100) & 51881                                                                   \\
		\end{tabular}
	\end{center}
\end{table}

\begin{table}[t]
	\caption{Summary of the graph datasets}
	\centering
	\begin{threeparttable}
		\begin{tabular}{llll}
			
			\textbf{CHARACTERISTIC} & \textbf{CORA} & \textbf{PUBMED} &\textbf{PPI} \\
			\hline
			Number of Vertices & 2708 & 19717 &  56944, 2373\tnote{a} \\
			Average Degree & 3.898 & 4.496 & 28.8\tnote{a} \\
			Number of Vertex Features & 1433 & 500 & 50 \\
			Number of Classes & 7 & 3 & 121\tnote{b} \\
			Number of Training Vertices & 1208 & 18217 & 44906\tnote{c} \\
			Number of Test Vertices & 1000 & 1000 & 5524\tnote{c}  \\
			
		\end{tabular}
		\begin{tablenotes}
			\small
			\item [a] The PPI dataset comprises several graphs, so the quantities marked with an ``a", represent the characteristic of the average graph .
			\item [b] For PPI, there are 121 targets, each taking values in $\{0, 1\}$.
			\item [c] All of the training nodes come from 20 graphs while the testing nodes come from two graphs not utilized during training.
		\end{tablenotes}
	\end{threeparttable}
	\label{tab:sumstats}
\end{table}

The number of trainable parameters in each model is independent of $k_1$ and $k_2$ by the design of LSTMs (the same is true for the mean-pooling aggregator).  The only variation in the number of weights is in the dimensions of the input and output features, which differ by dataset.  Please see Table~\ref{tab:NumParamsGraphs} for details.

Optimization is performed with the Adam optimizer~\citep{KingmaAdam}.  The training routine for the smaller graphs (Cora, Pubmed) is not guaranteed to see the entire training data, in contrast with the scheme applied to the larger PPI graph.  For Cora and Pubmed, we form 100 minibatches by randomly sampling subsets of 256 vertices from the training dataset (with replacement).  With PPI, we perform 10 full epochs: at each epoch, the training data is shuffled, partitioned into minibatches of size 512, and we pass over each.  In either case, the weights are updated after computing the gradient of the loss on each minibatch.

The hyperparameters were set by following~\citet{Hamilton2017}; no hyperparameter optimization was performed.  For every dataset, the embedding dimension was set to $q=256$ at both conv layers.  For Pubmed and PPI, the learning rate is set at 0.01 while for Cora it is set at 0.005.  

At test time, we load the weights obtained from training, perform 20 forward passes -- which shuffles the input sequence by design -- average the predicted probabilities (i.e. softmax output) from each forward pass, and choose the class that maximizes the averaged probabilities.  

The implementation for Mean Pooling is similar in spirit but replaces $\harrow{f}$ with a permutation invariant function.  The details can be found by viewing our repo on GitHub.

\begin{table}[t]
	\vspace{-5pt}
	\caption{Number of trainable parameters for each model in the graph task.  The number does not depend on $k_1$ or $k_2$.}
	\centering
	\vspace{-5pt}

	\begin{tabular}{llll}
		
		\emph{$\harrow{f}$} & \textbf{CORA} & \textbf{PUBMED} &\textbf{PPI} \\
		\hline
		mean-pool & 400512 & 161152 &  61056 \\
		LSTM      & 2541440 & 1465600 & 977408 \\

	\end{tabular}

	\label{tab:NumParamsGraphs}
\end{table}


\section{\LaTeX for \Janossy function markers}

The commands below can be directly copied and pasted into \LaTeX source to create the \Janossy function markers.  Please use the \texttt{amsmath} package.

To typeset $\dbar{f}$, we define in \LaTeX as
\begin{lstlisting}
\newcommand*\dbar[1]{\overline{\overline{\lower0.2ex\hbox{$#1$}}}}
\end{lstlisting}
and type
\begin{lstlisting}
$\dbar{f}$.
\end{lstlisting}
Similarly, for $\harrow{f}$, we define in \LaTeX as
\begin{lstlisting}
\newcommand{\harrow}[1]{\mathstrut\mkern2.5mu#1\mkern-11mu\raise1.6ex
\hbox{$\scriptscriptstyle\rightharpoonup$}}
\end{lstlisting}
and type
\begin{lstlisting}
$\harrow{f}$.
\end{lstlisting}

Last, the definition above for $\harrow{f}$ caused difficulties in environments such as \texttt{figure}, so we defined and occasionally used in \LaTeX
\begin{lstlisting}
\newcommand{\harrowStable}[1]{\overset{\rightharpoonup}{#1}}.
\end{lstlisting}

\ifdefined\COMPLETE
\else
\end{document}
\fi

\end{document}